\pdfoutput=1

\documentclass[11pt]{article}
\usepackage[preprint]{acl}

\usepackage{amsmath,amssymb,amsfonts}
\usepackage{graphicx}
\usepackage{textcomp}
\usepackage{xcolor}
\usepackage{pdfpages}
\usepackage{booktabs}
\usepackage{tabularx}
\usepackage{amsmath}
\usepackage{hyperref}
\usepackage{multirow}
\usepackage{array}
\usepackage{tcolorbox}
\usepackage{times}
\usepackage{latexsym}
\usepackage{multirow}
\usepackage{placeins}
\usepackage{longtable}
\usepackage{makecell}
\usepackage{amssymb}
\usepackage{etoolbox}
\usepackage{mdframed}
\usepackage{threeparttable}

\usepackage{siunitx}
\usepackage{colortbl}

\usepackage[T1]{fontenc}
\usepackage[utf8]{inputenc}
\usepackage{subcaption}
\usepackage{upquote}
\usepackage{inconsolata}
\usepackage{graphicx}
\usepackage{pifont}
\newcommand{\cmark}{\textcolor{darkgreen}{\ding{51}}}
\newcommand{\xmark}{\textcolor{red}{\ding{55}}}
\usepackage{algorithm}
\usepackage{algpseudocode}
\tcbset{
  colback=white,
  colframe=black,
  boxrule=0.5pt,
  arc=0pt,
  left=0pt,
  right=0pt,
  top=4pt,
  bottom=4pt
}
\usepackage{url}
\definecolor{darkgreen}{rgb}{0,0.5,0}
\definecolor{cardbg}{RGB}{249,249,251}
\definecolor{cardframe}{RGB}{230,230,235}
\definecolor{headfg}{RGB}{90,90,105}

\newcolumntype{Y}{>{\raggedright\arraybackslash}p{0.23\linewidth}}

\tcbset{
  bnmmluCard/.style={
    enhanced,
    colback=cardbg,
    colframe=cardframe,
    boxrule=0.4pt,
    arc=2mm,
    left=6pt,right=6pt,top=5pt,bottom=6pt,
  },
  bnmmluRow/.style={
    enhanced,
    colback=white,
    colframe=black!5,
    boxrule=0pt,
    left=2pt,right=2pt,top=3pt,bottom=4pt
  }
}

\usepackage{enumitem}

\usepackage{mathtools} 

\tcbset{colback=white, colframe=black!70, boxsep=4pt, left=6pt, right=6pt, top=6pt, bottom=6pt}

\allowdisplaybreaks[2] 

\usepackage{tablefootnote}
\usepackage{lipsum}

\title{\textsc{BnMMLU}:\\Measuring Massive Multitask Language Understanding in Bengali}

{\small
\author{
Saman Sarker Joy\textsuperscript{1} \and Swakkhar Shatabda\textsuperscript{2}\\
\textsuperscript{1}University of Malaya\\
\textsuperscript{2}BRAC University\\
\texttt{saman.sarker.joy@gmail.com}, \texttt{swakkhar.shatabda@bracu.ac.bd}
}
}

\begin{document}


\def\sbng{\bngviii}
\def\tbng{\bngvi}
\def\bng{\bngx}
\def\lbng{\bngxiv}
\def\Lbng{\bngxviii}
\def\LBng{\bngxxii}
\def\hbng{\bngxxv}
\def\Hbng{\bngxxx}
\def\sbns{\bnsviii}
\def\tbns{\bnsvi}
\def\bns{\bnsx}
\def\lbns{\bnsxiv}
\def\Lbns{\bnsxviii}
\def\LBns{\bnsxxii}
\def\hbns{\bnsxxv}
\def\Hbns{\bnsxxx}
\def\sbnw{\bnwviii}
\def\tbnw{\bnwvi}
\def\bnw{\bnwx}
\def\lbnw{\bnwxiv}
\def\Lbnw{\bnwxviii}
\def\LBnw{\bnwxxii}
\def\hbnw{\bnwxxv}
\def\Hbnw{\bnwxxx}



\font\bngv=bang10 scaled 500
\font\bngvi=bang10 scaled 600
\font\bngvii=bang10 scaled 700
\font\bngviii=bang10 scaled 800
\font\bngix=bang10 scaled 900
\font\bngx=bang10
\font\bngxi=bang10 scaled 1100
\font\bngxii=bang10 scaled 1200
\font\bngxiv=bang10 scaled 1400
\font\bngxviii=bang10 scaled 1800
\font\bngxxii=bang10 scaled 2200
\font\bngxxv=bang10 scaled 2500
\font\bngxxx=bang10 scaled 3000

\font\bnsv=bangsl10 scaled 500
\font\bnsvi=bangsl10 scaled 600
\font\bnsvii=bangsl10 scaled 700
\font\bnsviii=bangsl10 scaled 800
\font\bnsix=bangsl10 scaled 900
\font\bnsx=bangsl10
\font\bnsxi=bangsl10 scaled 1100
\font\bnsxii=bangsl10 scaled 1200
\font\bnsxiv=bangsl10 scaled 1400
\font\bnsxviii=bangsl10 scaled 1800
\font\bnsxxii=bangsl10 scaled 2200
\font\bnsxxv=bangsl10 scaled 2500
\font\bnsxxx=bangsl10 scaled 3000

\font\bnwv=bangwd10 scaled 500
\font\bnwvi=bangwd10 scaled 600
\font\bnwvii=bangwd10 scaled 700
\font\bnwviii=bangwd10 scaled 800
\font\bnwix=bangwd10 scaled 900
\font\bnwx=bangwd10
\font\bnwxi=bangwd10 scaled 1100
\font\bnwxii=bangwd10 scaled 1200
\font\bnwxiv=bangwd10 scaled 1400
\font\bnwxviii=bangwd10 scaled 1800
\font\bnwxxii=bangwd10 scaled 2200
\font\bnwxxv=bangwd10 scaled 2500
\font\bnwxxx=bangwd10 scaled 3000


\def\*#1*#2{o\null{#2}{#1}}

\def\d#1{\oalign{\smash{#1}\crcr\hidewidth{$\!$\rm.}\hidewidth}}

\def\sh#1{\setbox0=\hbox{#1}%
     \kern-.02em\copy0\kern-\wd0
     \kern.04em\copy0\kern-\wd0
     \kern-.02em\raise.0433em\box0 }

\maketitle
\begin{abstract}
Large-scale multitask benchmarks have driven rapid progress in language modeling, yet most emphasize high-resource languages such as English, leaving Bengali underrepresented. We present BnMMLU, a comprehensive benchmark for measuring massive multitask language understanding in Bengali. BnMMLU spans 41 domains across STEM, humanities, social sciences, and general knowledge, and contains 134,375 multiple-choice question–option pairs-the most extensive Bengali evaluation suite to date. The dataset preserves mathematical content via MathML, and includes BnMMLU-HARD, a compact subset constructed from questions most frequently missed by top systems to stress difficult cases. We benchmark 24 model variants across 11 LLM families, spanning
open-weights general/multilingual, Bengali-centric open-weights, and proprietary models, covering multiple parameter scales and instruction-tuned settings. We evaluate models under standardized protocols covering two prompting styles (Direct vs. Chain-of-Thought) and two context regimes (0-shot vs. 5-shot), reporting accuracy consistently across families. Our analysis highlights persistent gaps in reasoning and application skills and indicates sublinear returns to scale across model sizes. We release the dataset and evaluation templates to support rigorous, reproducible assessment of Bengali language understanding and to catalyze progress in multilingual NLP.
\end{abstract}

\begin{table*}[ht]
  \centering
  \small
  \setlength{\tabcolsep}{2pt}
  \renewcommand{\arraystretch}{1.2}
  \begin{tabular*}{1.0\textwidth}{@{\extracolsep{\fill}}l l c c c c}
    \toprule
    \textbf{Dataset} & \textbf{Format} & \textbf{\# Items} & \textbf{\# Subjects} & \textbf{Math} & \textbf{S:H:SS:O} \\
    \midrule
    \midrule
    BanglaQuAD \cite{rony2024banglaquadbengaliopendomainquestion}
      & Extractive & 30{,}808 & 14 & \xmark & 3:4:5:2 \\
    BanglaRQA \cite{ekram-etal-2022-banglarqa}
      & Extractive & 14{,}889 & 20 & \xmark & 1:2:2:3 \\
    BEnQA \cite{shafayat-etal-2024-benqa}
      & MCQ        &  5{,}161 & 5  & \cmark & 1:0:0:0 \\
    BLUCK \cite{kabir2025bluckbenchmarkdatasetbengali}
      & MCQ        &  2{,}366 & 23 & \xmark & 0:1637:729:0 \\
    NOIRBETTIK \cite{Aurpa2025NOIRBETTIK}
      & MCQ        &  5{,}215 & 8  & \xmark & 2:8:1:4 \\
    TituLM-Bangla MMLU \cite{nahin2025titullmsfamilybanglallms}
      & MCQ        & 87{,}869 & 11 & \xmark & 98:19:17:1 \\
    UDDIPOK \cite{AURPA2023108933}
      & Extractive &  3{,}636 & -- & \xmark & -- \\
    \midrule
    \textbf{BnMMLU}
      & \textbf{MCQ}
      & \textbf{134{,}375}
      & \textbf{41}
      & \textbf{\cmark}
      & \textbf{4:2:3:1} \\
    \bottomrule
  \end{tabular*}
  \caption{Comparison of prominent Bengali QA datasets. The table lists format (extractive vs.\ multiple choice), size (items and subjects), preservation of mathematical content (MathML), and proportional distribution across STEM, Humanities, Social Sciences and Others.}
  \label{tab:good-tab}
\end{table*}

\section{Introduction}\label{introduction}



The advancement of natural language processing (NLP) has been significantly driven by large-scale benchmarks that assess the capabilities of language models across various domains. Among these, the Massive Multitask Language Understanding (MMLU) \cite{hendrycks2021measuring} benchmark has emerged as a widely recognized evaluation framework. MMLU covers 57 diverse subjects, spanning disciplines such as mathematics, science, humanities, history, law, medicine and general knowledge. It is designed to measure a model’s ability to generalize across multiple domains. While MMLU has significantly contributed to evaluating models in high-resource languages like English, it provides little to no coverage for low-resource languages.

Although Bengali\footnote{We use \textit{Bengali} and \textit{Bangla} interchangeably to denote the same language (ISO 639-1: \texttt{bn}; ISO 639-3: \texttt{ben}). The IANA Language Subtag Registry entry for \texttt{bn} lists both (\url{https://www.iana.org/assignments/language-subtag-registry}).} is the seventh most spoken language globally \cite{statista_languages}, Bengali remains underrepresented in NLP research, with limited high-quality datasets, pre-trained models and benchmarks. The absence of a standardized knowledge-driven evaluation data set for Bengali language models restricts their ability to generalize across real-world tasks. While some multilingual benchmarks include Bengali \cite{kakwani-etal-2020-indicnlpsuite}, their coverage is sparse and does not adequately test subject-specific knowledge or reasoning skills in Bengali

In the absence of such a benchmark, researchers lack the means to assess whether a model’s responses in Bengali reflect genuine understanding, memorization of bilingual cues or hallucination. Our study is guided by the following:

\begin{enumerate}[label=(RQ\arabic*), leftmargin=*, itemsep=2pt]
  \item How far do multilingual vs.\ Bengali-centric models transfer to native Bengali tasks across various domains?
  \item What are the returns to scale under standardized prompting/context regimes?
  \item When does elicited reasoning help (or hurt), especially on difficult items?
  \item Which subject areas are systematically hard vs.\ easy across different LLMs?
\end{enumerate}

To address these questions, we introduce BnMMLU, a benchmark to evaluate the multitask language understanding of Bengali in language models. Our contributions in this work are:

\begin{itemize}[itemsep=2pt]
    \item A 41-domain MCQ suite with \textbf{134{,}375} spanning STEM, humanities, social science and general knowledge.
    \item Introduced \textsc{BnMMLU-HARD}, formed by ranking questions most frequently missed by models while preserving subdomain balance for stress testing.
    \item Evaluated 24 model variants, spanning open-weights general/multilingual, Bengali-centric open-weights and proprietary models.
    \item Comparable reporting across Direct vs. CoT and 0-shot vs. 5-shot settings and Reasoning and Non-reasoning comparisons with consistent prompts and accuracy metrics.
\end{itemize}

\begin{figure*}[ht]
  \centering
  \includegraphics[width=0.9\linewidth]{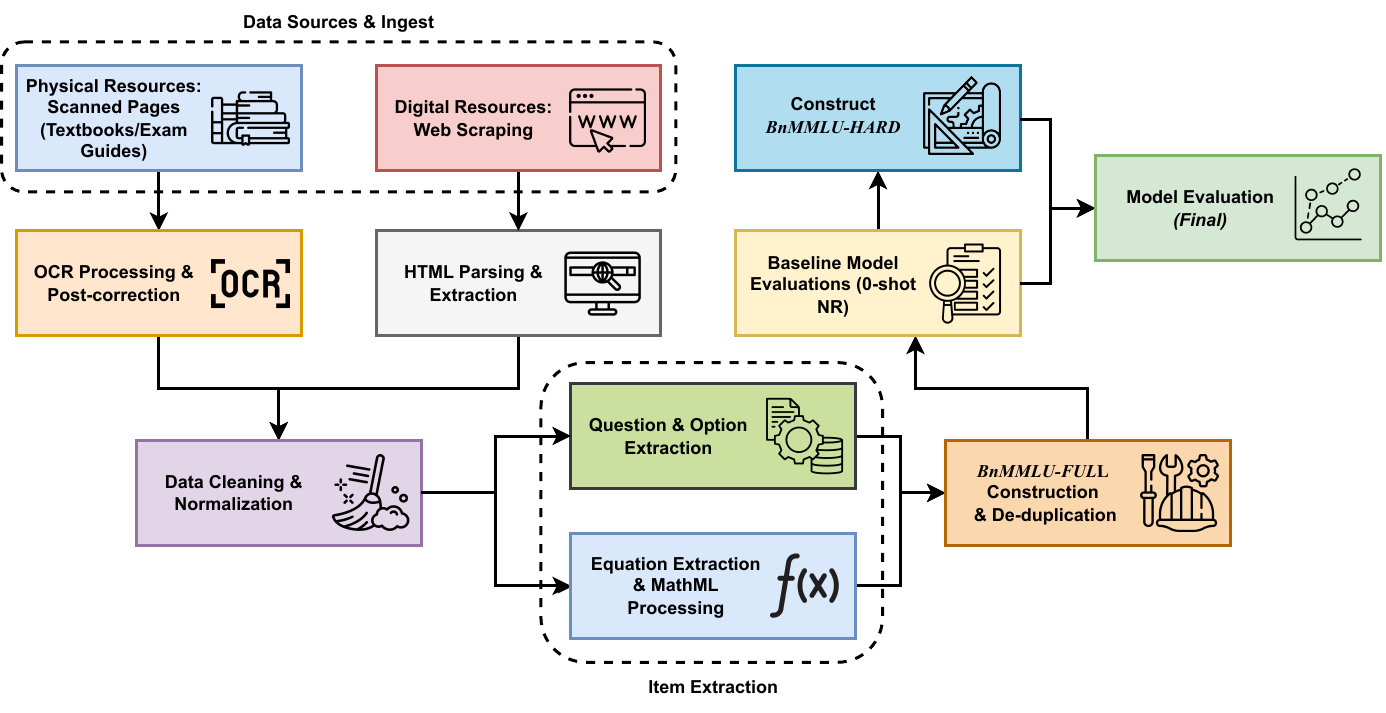} 
  \caption{An overview of the pipeline for constructing the BnMMLU benchmark.}
  \label{fig:example}
\end{figure*}

\section{Related Work}\label{literature-review}

The Massive Multitask Language Understanding (MMLU) benchmark \cite{hendrycks2021measuring} set a standard for evaluating language models on broad domain knowledge (e.g., mathematics, science, humanities, law), but is essentially English-centric and does not capture the linguistic, cultural and syntactic nuances of other languages.

Language-specific MMLU-style benchmarks extend this paradigm to local exams: KMMLU for Korean \cite{son2024kmmlumeasuringmassivemultitask}, CMMLU for Chinese \cite{li2024cmmlumeasuringmassivemultitask}, and ArabicMMLU for Modern Standard Arabic \cite{koto2024arabicmmluassessingmassivemultitask}, all reporting that non-English models still lag behind their English counterparts.

In the multilingual setting, IndicGLUE \cite{kakwani-etal-2020-indicnlpsuite} and XGLUE \cite{liang-etal-2020-xglue} include Bengali among many languages and cover tasks such as classification, sentiment analysis, NER and QA, but they are not broad multitask knowledge benchmarks in the MMLU sense.

For Bengali specifically, existing resources such as BanglaQuAD \cite{rony2024banglaquadbengaliopendomainquestion}, BanglaRQA \cite{ekram-etal-2022-banglarqa}, BEnQA \cite{shafayat-etal-2024-benqa}, and related datasets (e.g., NOIRBETTIK, BLUCK) provide task-specific QA and reading-comprehension style evaluations, often focusing on span extraction or short-answer questions rather than structured, curriculum-style subject coverage.

TituLM-Bangla MMLU \cite{nahin2025titullmsfamilybanglallms} adapts MMLU-style diagnostics to Bengali multiple-choice questions across various topics, but with narrower subject breadth and less fine-grained coverage than our \textsc{BnMMLU}, which targets a wider set of Bengali academic and professional domains for multitask knowledge and reasoning evaluation.

\section{The BnMMLU Benchmark}\label{theoretical-framework}
We create BnMMLU, a multitask benchmark composed of multiple-choice question–answer pairs across 41 subjects spanning STEM, humanities, social sciences and other domains. We refer to this complete benchmark as \textbf{BnMMLU-FULL} throughout the remainder of the paper. The overview of the full pipeline is shown in \autoref{fig:example}.

\subsection{Dataset Construction}
The questions were sourced from Bangladeshi educational and professional materials through two channels.

\paragraph{Physical Resources.} Scanned pages from NCTB-approved textbooks and competitive exam guides, processed using OCR tool with post-correction for script accuracy. Due to the unstructured formatting of many print materials, 20\% of the data came from these sources, and they did not contain properly formatted multiple-choice questions and answers. Examples of these books are shown in \autoref{fig:three_with_borders}.
    
\paragraph{Digital Sources.} \label{sec:dataset-construction} Web-scraped questions from Bangladeshi educational portals that host structured, exam-style multiple-choice questions. The web scraping was performed using Selenium\footnote{\href{https://www.selenium.dev/}{https://www.selenium.dev}} and BeautifulSoup\footnote{\href{https://pypi.org/project/beautifulsoup4/}{https://pypi.org/project/beautifulsoup4}}. The majority of the dataset, around 80\% of the data came from these digital sources.

\subsection{Optical Character Recognition (OCR) \& Post-Correction} \label{subsec:ocr}
We scan printed book pages, apply a standard pre-processing pipeline (grayscale conversion, adaptive binarisation, and deskewing) and then run OCR system followed by LLM-based copy-editing to clean the text while preserving math and answer keys. Full implementation details and the exact copy-editing prompt are provided in \autoref{sec:ocr-details}.

Post-correction reduced formatting issues and spelling errors. Additionally, approximately 10\% of the question-option pairs were manually reviewed by the authors to ensure that the OCR text, math expressions and answer keys matched the original source pages.



\begin{table*}[h]
\centering
\small
\setlength{\tabcolsep}{4pt}
\renewcommand{\arraystretch}{1.2}
\begin{tabular}{lccccc}
\toprule
\textbf{Model (Best per Family)} &
\textbf{STEM} &
\textbf{Humanities} &
\textbf{Social Sciences} &
\textbf{Others} &
\textbf{Overall ($\Delta$)} \\
\midrule
\midrule
\multicolumn{6}{c}{\textit{\textbf{English-Centric / Bilingual Instruction-Tuned Models (Best per Family)}}} \\
\midrule
\textsc{Llama-3.3-70B-Instruct}     & 62.53 & 52.47 & \underline{65.81} & \underline{68.99} & 61.87 \textcolor{darkgreen}{(+36.87)} \\
\textsc{Qwen3-32B}                  & \underline{72.03} & \underline{53.01} & 65.57 & 66.44 & \underline{65.34 \textcolor{darkgreen}{(+40.34)}} \\
\textsc{Gemma-3-27B-IT}             & 63.61 & 51.43 & 63.90 & 67.44 & 61.27 \textcolor{darkgreen}{(+36.27)} \\
\midrule
\multicolumn{6}{c}{\textit{\textbf{Bengali Pretrained / Instruction-Tuned Models (Best per Family)}}} \\
\midrule
\textsc{TituLLM-1B}                 & 27.15 & 27.53 & 28.42 & 28.19 & 27.72 \textcolor{darkgreen}{(+2.72)} \\
\textsc{TigerLLM-9B-IT}             & \underline{56.02} & \underline{47.85} & \underline{59.48} & \underline{61.29} & \underline{55.70 \textcolor{darkgreen}{(+30.70)}} \\
\textsc{BanglaLLaMA-3.1-8B-Instruct} & 26.10 & 27.56 & 27.58 & 26.52 & 26.95 \textcolor{darkgreen}{(+1.95)} \\
\midrule
\multicolumn{6}{c}{\textit{\textbf{Proprietary Models}}} \\
\midrule
\textsc{GPT-5-Mini}                 & 48.25 & 43.96 & 55.00 & 55.78 & 50.09 \textcolor{darkgreen}{(+25.09)} \\
\textsc{Grok 4 Fast}                & 61.98 & 51.63 & 64.02 & 67.60 & 60.82 \textcolor{darkgreen}{(+35.82)} \\
\textsc{Gemini 2.5 Flash}      & 72.38 & \textbf{62.32} & \textbf{71.08} & \textbf{73.85} & \textbf{69.85 \textcolor{darkgreen}{(+44.85)}} \\
\textsc{DeepSeek-V3.2-Exp}          & 72.72 & 58.62 & 70.06 & 73.84 & 68.82 \textcolor{darkgreen}{(+43.82)} \\
\textsc{Qwen-Plus}                  & \textbf{73.49} & 56.15 & 66.89 & 70.17 & 67.29 \textcolor{darkgreen}{(+42.29)} \\
\bottomrule
\end{tabular}
\caption{Average accuracy (\%) of models on the \textbf{BnMMLU-FULL} benchmark under 0-shot Direct (Non-Reasoning) evaluation. We report only the best-performing checkpoint per model family. Bold marks the highest overall score; underlines denote the best model within each category. ($\Delta$) in overall is compared with random baseline (25\%).}
\label{tab:bnmmlu_full_best_family}
\end{table*}

We stored all equations in MathML\footnote{\href{https://www.w3.org/TR/MathML/}{https://www.w3.org/TR/MathML}}, an XML-based markup language for representing mathematics.

\subsection{Duplicate-Question De-duplication}
We embed each question–option string using \textit{text-embedding-3-small}\footnote{\href{https://platform.openai.com/docs/models/text-embedding-3-small}{https://platform.openai.com/docs/models/text-embedding-3-small}} and run approximate nearest-neighbor search with the angular metric. For each question $q_i$, we retrieve top-$k$ neighbors and convert angular distances $d(q_i,q_j)$ to similarities $s(q_i,q_j) = 1 - d(q_i,q_j)/2$; pairs with $s \ge 0.90$ define edges in an undirected graph whose connected components form duplicate clusters. We keep a single canonical item per cluster to obtain a de-duplicated benchmark. Full details are in \autoref{app:duplicate}.

\subsection{Task Categories}
The benchmark covers 41 subjects across STEM, Humanities, Social Sciences and Other domains; a full list of subjects and tested concepts is provided in \autoref{sec:subject_list}.

\subsection{Training-test decontamination}
\label{sec:decontamination}

Because roughly 80\% of \textsc{BnMMLU} is sourced from web-based question banks (\autoref{sec:dataset-construction}), we explicitly quantify potential train-test overlap on LLMs via an $n$-gram decontamination analysis. Following the GPT-3 contamination protocol and subsequent work, a shared 13-token span is treated as a conservative signal of near-verbatim memorization rather than chance overlap \citep{10.5555/3495724.3495883, ravaut2025comprehensivesurveycontaminationdetection}.

Overall contamination is low: for most corpora, fewer than $0.1\%$ of questions exhibit any overlapping 13-gram. Full preprocessing details, per-corpus breakdowns are provided in \autoref{app:decontamination}.

\subsection{BnMMLU-HARD}
\label{subsec:BnMMLU-HARD}
We construct \textbf{BnMMLU-HARD} as a compact subset focused on the questions most frequently missed by the \textit{top-10} models on \textbf{BnMMLU-FULL}, using their 0-shot (Direct) scores. Questions are ranked by aggregate error across these models, and we select the highest-error set while preserving a proportional subdomain balance. The distribution for both of them is shown in \autoref{fig:placeholder}.

\section{Experimental Evaluation}\label{research-methodology}

Following the recommendation from prior work \cite{lai2023chatgptenglishcomprehensiveevaluation}, we keep the system prompt in English unless stated otherwise. 


\begin{table*}[h]
\centering
\small
\setlength{\tabcolsep}{4pt}
\renewcommand{\arraystretch}{1.2}
\begin{tabular}{lcccc}
\toprule
\textbf{Model} &
\makecell{\textbf{0-shot Direct}\\\textbf{(Non-Reasoning)}} &
\makecell{\textbf{0-shot CoT ($\Delta$)}\\\textbf{(Non-Reasoning)}} &
\makecell{\textbf{5-shot Direct}\\\textbf{(Non-Reasoning)}} &
\makecell{\textbf{5-shot CoT ($\Delta$)}\\\textbf{(Non-Reasoning)}} \\
\midrule
\midrule
\multicolumn{5}{c}{\textit{\textbf{English-Centric / Bilingual Instruction-Tuned Models}}} \\
\midrule
\textsc{Llama-3.2-3B-Instruct}            & 19.95 & 18.33 \textcolor{red}{(-1.62)} & 22.16 & 23.25 \textcolor{darkgreen}{(+1.09)} \\
\textsc{Llama-3.3-70B-Instruct}           & 23.78 & 35.17 \textcolor{darkgreen}{(+11.39)} & 31.15 & \underline{37.50 \textcolor{darkgreen}{(+6.35)}} \\
\textsc{Qwen3-14B}                        & 14.67 & 14.32 \textcolor{red}{(-0.35)} & 18.35 & 16.88 \textcolor{red}{(-1.47)} \\
\textsc{Qwen3-32B}                        & \underline{25.52} & 28.63 \textcolor{darkgreen}{(+3.11)} & 34.63 & 31.19 \textcolor{red}{(-3.44)} \\
\textsc{Gemma-3-12B-IT}                   & 10.54 & 14.55 \textcolor{darkgreen}{(+4.01)} & 18.50 & 23.52 \textcolor{darkgreen}{(+5.02)} \\
\textsc{Gemma-3-27B-IT}                   & 14.72 & \underline{37.59 \textcolor{darkgreen}{(+22.87)}} & \underline{35.65} & 34.65 \textcolor{red}{(-1.00)} \\
\midrule
\multicolumn{5}{c}{\textit{\textbf{Bengali Pretrained / Instruction-Tuned Models}}} \\
\midrule
\textsc{TigerLLM-9B-IT}                   & \underline{11.01} & \underline{16.78 \textcolor{darkgreen}{(+5.77)}} & \underline{18.44} & \underline{23.32 \textcolor{darkgreen}{(+4.88)}} \\
\midrule
\multicolumn{5}{c}{\textit{\textbf{Proprietary Models}}} \\
\midrule
\textsc{GPT-5-Mini}                       & 14.13 & 19.12 \textcolor{darkgreen}{(+4.99)} & 19.66 & 18.63 \textcolor{red}{(-1.03)} \\
\textsc{Grok 4 Fast}                      & 20.94 & 20.89 \textcolor{red}{(-0.05)} & 44.06 & 51.12 \textcolor{darkgreen}{(+7.06)} \\
\textsc{Gemini 2.5 Flash}            & \textbf{34.46} & 45.38 \textcolor{darkgreen}{(+10.92)} & 51.62 & 61.48 \textcolor{darkgreen}{(+9.86)} \\
\textsc{DeepSeek-V3.2-Exp}                & 29.89 & \textbf{59.04 \textcolor{darkgreen}{(+29.15)}} & \textbf{58.83} & \textbf{64.53 \textcolor{darkgreen}{(+5.70)}} \\
\textsc{Qwen-Plus}                        & 32.47 & 58.74 \textcolor{darkgreen}{(+26.27)} & 57.40 & 55.09 \textcolor{red}{(-2.31)} \\
\bottomrule
\end{tabular}
\caption{Accuracy (\%) on \textbf{BnMMLU-HARD} for a reduced set of representative models. \(\Delta\) is computed as \(\text{CoT} - \text{Direct}\) at the \emph{same shot} (0-shot or 5-shot). \textbf{Bold} marks the global best per column; \underline{underline} marks the best \emph{within each category} per column.}
\label{tab:bnmmlu_3col_subdelta}
\end{table*}

\subsection{Model Selection}
We evaluate a diverse set of language models on the \textbf{BnMMLU} dataset. Our selection is designed to cover both proprietary and open-weight families, multiple parameter scales, instruction-tuned checkpoints where available and a balance between Bengali-centric and English-centric models. Detailed access and setup information is provided in \autoref{tab:model_overview_grouped}.

\subsection{Evaluation Protocol}
We evaluate each model under two prompting styles (Direct and Chain-of-Thought, CoT), two context regimes (0-shot and 5-shot) and two reasoning configurations (Reasoning-On and Non-Reasoning).

\paragraph{Exemplar Construction for 5-shot.}
We selected five questions from each domain and used \textsc{GPT-5-Mini} WebUI\footnote{\url{https://chatgpt.com/}} to make reasoning traces (CoT) the prompt in \autoref{fig:cot-example}. Then we manually screened the exemplars for correctness and style consistency. These were used as in-context demonstrations in the 5-shot setting (Direct uses the same exemplars but with the reasoning text removed).

\subsection{Evaluation Metrics} For evaluating performance on \textbf{BnMMLU-FULL} \& \textbf{BnMMLU-HARD}, we use accuracy as the primary metric. Accuracy is defined as the proportion of correctly predicted answers out of the total questions attempted.

\section{Discussion}\label{error-analysis}


\autoref{tab:bnmmlu_full_best_family} summarizes 0-shot Direct (Non-Reasoning) accuracy on \textbf{BnMMLU-FULL} and detailed summary is shown in \autoref{tab:bnmmlu_full}. Proprietary models lead overall: \textsc{Gemini 2.5 Flash} tops the chart (69.85) with best or near-best scores across Humanities, Social Sciences, and Others, while \textsc{Qwen-Plus} holds the STEM peak (73.49) and strong overall (67.29). Among open-weights, \textsc{Qwen3-32B} (65.34) and \textsc{Llama-3.3-70B-Instruct} (61.87) are the strongest, followed closely by \textsc{Gemma-3-27B-IT} (61.27).

Bengali-centric models show competitive mid-tier performance led by \textsc{TigerLLM-9B-IT} (55.70; best in its group), while small Bengali models cluster near the high-20s. Domain-wise, STEM tends to be the highest-scoring slice for top systems, with Humanities relatively lower for open-weights. Net: proprietary models currently set the frontier, large open-weights close much of the gap, and targeted Bengali pretraining helps at moderate scale but has not yet matched the largest bilingual/global families.

\begin{table*}[h]
\centering
\begingroup
\small
\setlength{\tabcolsep}{4pt}
\renewcommand{\arraystretch}{1.2}
\begin{tabular*}{0.9\textwidth}{@{\extracolsep{\fill}}lcccccccccc@{}}
\toprule
\multirow{2}{*}{\textbf{Model}} &
\multicolumn{2}{c}{\textbf{STEM}} &
\multicolumn{2}{c}{\textbf{Humanities}} &
\multicolumn{2}{c}{\textbf{Social Sciences}} &
\multicolumn{2}{c}{\textbf{Others}} &
\multicolumn{2}{c}{\textbf{Overall}} \\
\cmidrule(lr){2-3}\cmidrule(lr){4-5}\cmidrule(lr){6-7}\cmidrule(lr){8-9}\cmidrule(lr){10-11}
& NR & R & NR & R & NR & R & NR & R & NR & R \\
\midrule
\midrule
\textsc{Qwen3-32B}    & 35.41 & 68.76 & 13.07 & 27.82 & 20.24 & 37.12 & 20.02 & 41.57 & 25.52 & 49.41 \\
\textsc{GPT-5-Mini}   & 14.88 & 69.51 & 10.57 & 33.29 & 14.73 & 47.15 & 15.04 & 59.20 & 14.13 & 55.25 \\
\textsc{Grok 4 Fast}  & 22.12 & 77.34 & 15.68 & 44.79 & 21.10 & 57.61 & 25.38 & 68.01 & 20.94 & 64.64 \\
\textsc{Gemini 2.5 Flash} & 37.62 & 73.75 & 26.86 & 47.45 & \textbf{33.14} & 57.18 & \textbf{39.56} & 67.91 & \textbf{34.46} & 63.39 \\
\textsc{DeepSeek-V3.2-Exp} & 35.17 & \textbf{80.65} & \textbf{28.11} & \textbf{51.83} & 27.00 & \textbf{63.90} & 29.00 & \textbf{74.43} & 29.89 & \textbf{69.79} \\
\textsc{Qwen-Plus}    & \textbf{43.65} & 77.93 & 19.74 & 46.24 & 25.47 & 56.89 & 28.16 & 67.15 & 32.47 & 64.83 \\
\bottomrule
\end{tabular*}
\endgroup
\caption{0-shot Direct evaluation accuracy (\%) of reasoning-capable models on the \textbf{BnMMLU-HARD} subset. \textbf{NR} denotes Non-Reasoning and \textbf{R} denotes Reasoning.}
\label{tab:bnmmlu_hard_reasoning_models}
\end{table*}



\begin{figure*}[h]
\centering
\includegraphics[width=0.9\linewidth]{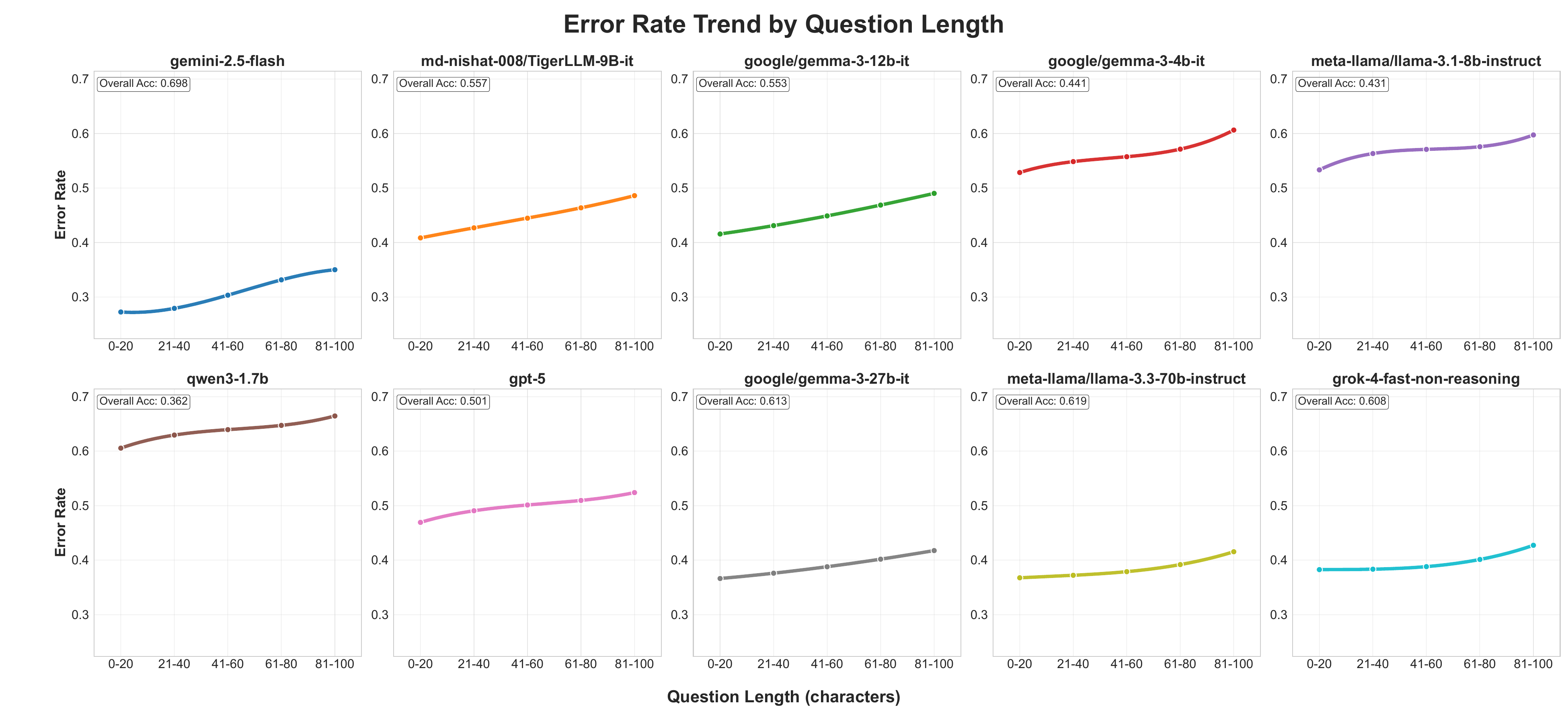}
\caption{Error rate trends by question length (in characters) across ten evaluated models on the \textbf{BnMMLU-FULL} benchmark. Each subplot represents an individual model, with the x-axis indicating question length bins and the y-axis showing corresponding error rates. Overall accuracy for each model is annotated in its respective panel for reference.}

\label{fig:graph_ref}
\end{figure*}

So, scale helps but with diminishing returns; consistent ladders imply healthy training pipelines; and matched-compute gaps highlight the outsized role of data and recipe design, especially beyond the mid-compute regime.

\subsection{Prompting \& Context Regimes}
As shown in \autoref{tab:bnmmlu_3col_subdelta}, adding reasoning and shots generally boosts accuracy, with the largest gains typically from \emph{5-shot CoT}. Standout jumps include \textsc{DeepSeek-V3.2-Exp} (29.89→\textbf{64.53}, +34.64), \textsc{Gemini 2.5 Flash} (34.46→61.48, +27.02), and \textsc{Grok 4 Fast} (20.94→51.12, +30.18). Among open weights, \textsc{Gemma-3-27B-IT} benefits markedly (+22.87 with 0-shot CoT; +20.93 with 5-shot Direct), and \textsc{Llama-3.3-70B-Instruct} rises to 37.50 (+13.72). Bengali-centric \textsc{TigerLLM-9B-IT} starts low (11.01) but more than doubles under 5-shot CoT (23.32; +12.31), indicating prompting can partly offset limited scale.

Gains are heterogeneous and sometimes negative at small–mid scales: \textsc{Llama-3.2-3B-Instruct} (0-shot CoT: -1.62), \textsc{Qwen3-8B} (-0.68), and \textsc{Qwen3-14B} (-0.35); moreover, 5-shot CoT can underperform 5-shot Direct in some cases (e.g., \textsc{Qwen3-32B}: +5.67 vs.\ +9.11).

\subsection{Reasoning Effects}
Across all reasoning-capable models on BnMMLU-HARD, enabling reasoning consistently lifts accuracy in every domain and for every model. Overall gains range from \textsc{Qwen3-1.7B} (+14.75; 14.53$\rightarrow$29.28) to \textsc{Grok 4 Fast} (+43.70; 20.94$\rightarrow$64.64), with substantial jumps also for \textsc{GPT-5-Mini} (+41.12) and \textsc{DeepSeek-V3.2-Exp} (+39.90). Under the reasoning setting, \textsc{DeepSeek-V3.2-Exp} attains the top scores across all domains-STEM 80.65, Humanities 51.83, Social Sciences 63.90, Others 74.43-and the highest overall (69.79). By contrast, under non-reasoning, the strongest baselines are split: \textsc{Qwen-Plus} leads STEM (43.65), \textsc{DeepSeek-V3.2-Exp} leads Humanities (28.11), and \textsc{Gemini 2.5 Flash} leads Social Sciences (33.14), Others (39.56), and Overall (34.46). These patterns indicate that reasoning particularly amplifies STEM and “Others” performance for mid/large models (e.g., \textsc{Qwen3-14B} STEM 18.42$\rightarrow$65.36; \textsc{Grok 4 Fast} Others 25.38$\rightarrow$68.01), while still yielding reliable improvements in Humanities and Social Sciences. All figures are on \autoref{tab:bnmmlu_hard_reasoning_models}.

\subsection{Sequence-Length Robustness}

\begin{figure*}[h]
\centering
\includegraphics[width=0.9\linewidth]{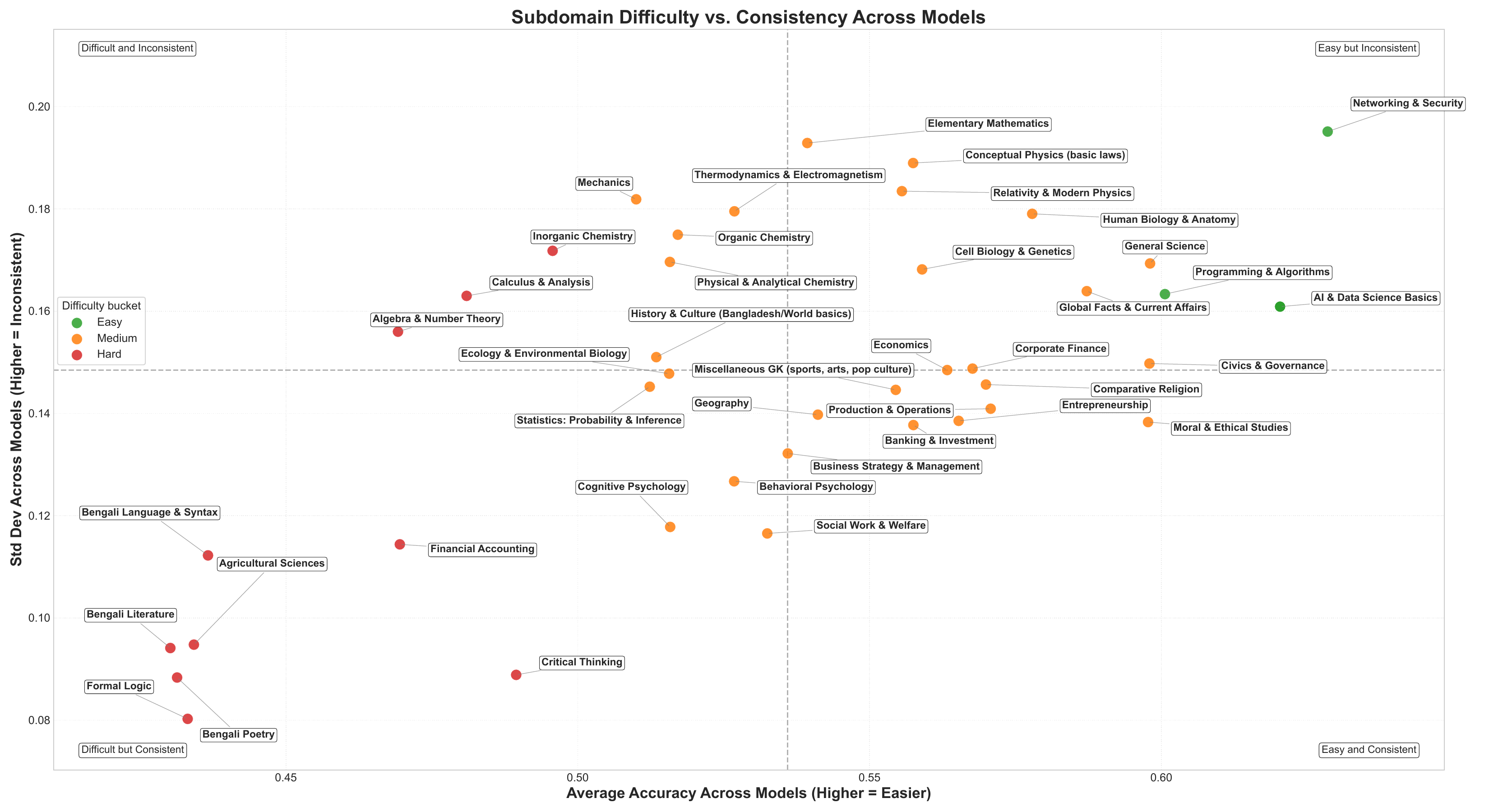}
\caption{Subdomain difficulty versus cross-model consistency on the \textbf{BnMMLU-FULL} benchmark under 0-shot Direct prompting. The x-axis shows mean accuracy across models (higher = easier), and the y-axis shows standard deviation (higher = more inconsistent); each point is a subdomain color-coded by difficulty bucket (\textit{Easy}, \textit{Medium}, \textit{Hard}). The four quadrants (\textit{Easy \& Consistent}, \textit{Easy but Inconsistent}, \textit{Difficult and Inconsistent}, \textit{Difficult but Consistent}) summarize how subdomain complexity and variability interact in assessing LLM robustness.}

\label{fig:fig1}
\end{figure*}


Across models, error rates increase monotonically with question length, with the sharpest degradation typically occurring between the $0$–$20$ and $81$–$100$ character bins. The strongest systems maintain the lowest error curves throughout: \textsc{gemini-2.5-flash}, \textsc{llama-3.3-70b-instruct} and \textsc{gemma-3-27b-it} show relatively shallow slopes as length grows. Mid-tier models such as \textsc{gemma-3-12b-it}, \textsc{TigerLLM-9B-it}, \textsc{GPT-5-Mini} exhibit a clearer length penalty past $60$ characters. Smaller/earlier-generation instruction models like \textsc{llama-3.1-8b-instruct}, \textsc{gemma-3-4b-it} and \textsc{qwen3-1.7b} have the highest error rates and the steepest length-dependent drop-offs. Consequently, the performance gap between top and weaker models widens in the longest bin, indicating reduced robustness to longer, likely more compositionally complex, prompts. The per-model length-specific error profiles are visualised in \autoref{fig:graph_ref}.






\subsection{Subject-Specific Failure Modes}
\paragraph{Analysis.}
\label{sec:subject-failure}
Using median-based splits (dashed lines), we see four bands: Difficult \& Inconsistent-advanced STEM (e.g., algebra/analysis, inorganic chemistry, mechanics) with low accuracy and wide spread; 
Easy \& Inconsistent-computing/tech survey areas (networking, AI/data basics, programming) that score high but vary by model; 
Difficult \& Consistent-Bengali/logic plus applied topics (accounting, agriculture) that are uniformly hard; 
Easy \& Consistent-management/psych/finance/geography that most models handle reliably. \autoref{fig:fig1} shows the domain difficulty versus consistency.

\subsection{Error Taxonomy \& Case Studies}
\label{subsec:error-taxonomy}

Across all prompting regimes, we observe a small set of recurring slips that surface with different facades.

\paragraph{Instruction-Following vs. Heuristic Shortcuts.}


A first class of errors stems from the model seizing a plausible heuristic instead of following the full instruction. When asked to expand an acronym, for instance, the model often latches onto the most \textit{frequent} completion rather than the \textit{domain-correct} one. In one item (``e-GP stands for: \ldots''), a 0-shot direct answer defaulted to \textbf{electronic government purchasing}, likely because ``purchasing'' is a frequent neighbor of ``e-government'' in pretraining text. Chain-of-thought (CoT) prompting nudged the model to reason about procurement systems and public-sector terminology, which shifted the answer to \textbf{electronic government procurement} - the intended domain term. CoT slows the jump to a high-frequency collocation and creates space to align to the task's governing instruction (disambiguate by function, not by frequency).

\paragraph{Ambiguity at the Interface: Formatting, Scripts, and Mixed Notation.}

A second cluster originates \textbf{upstream of reasoning}: mixed scripts (Bengali + Roman), MathML-like tokens (\verb|<msup>|, \verb|<msqrt>|), and lookalike glyphs (``ln'' vs. ``1n'') can be partially misparsed, leading the model to answer a \emph{nearby} question.

\paragraph{Calibration and the Amount of Examples.}

More examples is not always better. We see overfitting to few-shot context where long Chain-of-Thought (CoT) rationales import the wrong frame (e.g., shown in \autoref{fig:bilashi}). Conversely, \textbf{right-sized scaffolds} 2-4 concise checks tied to the item’s gating cues (time unit, regime, exponent, unit) - deliver the largest flips from wrong to right.

\paragraph{Design Implications (Across Domains).}
Three prompt-level nudges generalize: (i) \emph{normalize then solve} for mixed markup/symbols; (ii) \emph{scaffold lightly} to surface intermediate commitments without inducing spurious patterns; and (iii) \emph{option-calibrate} by matching the derived condition (unit/exponent/scope) to the exact wording of the alternatives.


\subsection{Bengali-Specific Error Patterns}

Bengali items add characteristic frictions that interact with the taxonomy above. Below are several case studies illustrating these patterns.

\paragraph{Orthography \& Mixed Markup at the Math/Language Boundary.}

Bengali prose often co-occurs with inline MathML-style tags in the options. Under 0-shot direct prompting, models sometimes select the most \textbf{salient-looking option} (e.g., a tidy fraction or exponent) without fully parsing the markup. For instance, in questions involving calculus or optics, performance improves once the expression is restated in standard math and only then compared against candidates.

\paragraph{Anglicized Cue Phrases Inside Bengali Questions.}

Embedded English slogans or titles can bias frequency-driven guesses in direct mode. A single reasoning step that maps the phrase to world knowledge before selecting the option reliably corrects this.

\paragraph{Bengali Numerals, Currency Tokens and the Danda.}

Arithmetic questions mixing Bengali numerals with the word for currency (“Taka”) and closing with the Bengali danda tend to elicit rounded, visually salient choices in direct mode. Light reasoning that ties numerals to operations and checks the \textbf{unit phrase} flips such items to the correct answer.

\subsection{CoT vs. Reasoning-On}
\label{subsec:reasoning-case-report}

This section examines cases where \emph{5-shot CoT (non-reasoning)} answered incorrectly but \emph{0-shot Reasoning-On} answered correctly. We preserve the provided snippets and bold the decisive cues.


\paragraph{Why Reasoning-On Helps.}
Across slices where 5-shot CoT (non-reasoning) fails but 0-shot Reasoning-On succeeds, the dominant pattern is regime selection vs.\ heuristic lock-in. CoT often stabilizes on a salient rule and never revisits it; e.g., from \autoref{fig:physics} we can see, applying the inverse-square law outside Earth even though the query’s span is center $\rightarrow$ surface, where $g\!\propto\!r$ and thus increases proportionally (Mechanics; B). By contrast, Reasoning-On explicitly enumerates alternatives, chooses the inside-sphere regime, and then maps the wording to the option. A second failure mode is granularity misread: CoT carries over exemplar priors shown on \autoref{fig:gran}, about monthly limits and answers “4,” whereas Reasoning-On re-parses the temporal cue (per week) and validates against the weekly constraint before selecting “2” (Business Strategy \& Management; A). More generally, Reasoning-On performs lightweight option-checking after resolving the operative cue (temporal/categorical/physical), which prevents near-miss mappings and salience/round-number bias. The net effect is not longer chains, but earlier branching to the correct regime and a final consistency check with the provided options.

\section{Conclusion}\label{conclusion}

We introduced \textsc{BnMMLU}, a 41-subject Bengali benchmark of 134{,}375 multiple-choice questions spanning STEM, Humanities, Social Sciences, and Others, and \textsc{BnMMLU-HARD}, a stress-test subset constructed from items that strong systems most often miss. To support faithful Bengali evaluation, we preserve mathematical content (via MathML), normalize OCR-derived text, and apply de-duplication and training-test decontamination analyses. We benchmark a broad set of proprietary and open-weight LLMs under a controlled protocol covering prompting style (Direct vs.\ CoT), context regime (0-shot vs.\ 5-shot), and explicit reasoning configurations. Results show that proprietary systems still lead overall, while the best open-weight models narrow the gap; gains are largest when reasoning is enabled, especially on \textsc{BnMMLU-HARD}. Scaling trends generally improve accuracy but exhibit diminishing returns and meaningful cross-family differences, suggesting that data and post-training recipe quality matter beyond parameter count. Finally, we analyze robustness to question length and subject-specific failure modes to highlight where current models remain brittle. We release the benchmark and evaluation artifacts to enable reproducible measurement and to accelerate progress on Bengali language understanding and reasoning.

\section*{Limitations}
We evaluate text-only capabilities and do not cover multimodal settings (vision-aided reasoning), so the results may not reflect performance in real-world multimodal use cases. While we tested a broad set of models, we were constrained by compute and access costs; therefore, some newer, larger or more expensive frontier models (and larger-scale tuning/inference setups) were not included, which could shift absolute performance levels-though the benchmark remains useful for comparing models under a consistent, reproducible text-only evaluation setup.

\section*{Ethical Statement}\label{ethical-statement}
The dataset is publicly available under the CC BY-SA 4.0 license, ensuring free accessibility.

\section*{Conflicts of Interest}\label{conflicts-of-interest}

The authors declare that they have no conflicts of interest to this
work.

\section*{Acknowledgment}


No external funding was received.



\bibliography{custom}

\appendix


\newpage

\section{Task Categories}\label{sec:subject_list}

The task types include a broad range of academic and professional topics, each addressing a specific domain of expertise and practice. The subject list and its tested concepts are in \autoref{tab:mmlu_subjects}.

\paragraph{Humanities.}
Focuses on language, literature, philosophy, and ethics. Core areas include Bengali language \& syntax, Bengali literature and poetry, formal logic and critical thinking, comparative religion, and moral \& ethical studies. Coverage balances textual analysis with argumentation and value-oriented topics.

\paragraph{STEM (Science, Technology, Engineering, Mathematics).}
Emphasizes quantitative reasoning, natural sciences, and computing. Mathematics spans elementary topics, algebra \& number theory, calculus \& analysis, and statistics: probability \& inference; physics includes mechanics, thermodynamics \& electromagnetism, conceptual physics (basic laws), and relativity \& modern physics; chemistry covers physical \& analytical, inorganic, and organic subfields; life sciences include cell biology \& genetics, human biology \& anatomy, and ecology \& environmental biology. Computing tracks programming \& algorithms, networking \& security, AI \& data science basics, plus general science integration.

\paragraph{Social Sciences.}
Covers institutions, markets, and human behavior. Economics, banking \& investment, financial accounting, and corporate finance sit alongside business strategy \& management, production \& operations, and entrepreneurship. The domain also includes civics \& governance, geography, history \& culture, cognitive and behavioral psychology, and social work \& welfare.

\paragraph{Others.}
Includes general knowledge and global/current affairs, ranging from sports, arts, and media to international organizations, events, and world politics. Coverage reflects publicly available sources up to September 2024.

\section{OCR \& Post-Correction Details}
\label{sec:ocr-details}
Printed book pages were scanned at 300\,dpi into lossless TIFF images. Example scanned pages are shown in \autoref{fig:three_with_borders}. Then these images were pre-processed via (i) grayscale conversion, (ii) Sauvola adaptive thresholding, and (iii) Hough-transform deskewing before text extraction. We then employed \textsc{EasyOCR (v1.7.1)}\footnote{\href{https://github.com/JaidedAI/EasyOCR}{https://github.com/JaidedAI/EasyOCR}} with its Bengali language model to obtain raw transcriptions. OCR output was cleaned and formatted using \textsc{GPT-3.5-Turbo-0125}\footnote{\url{https://platform.openai.com/docs/models/GPT-3.5-Turbo}} via the OpenAI API, with the Bengali copy-editing prompt shown in \autoref{fig:prompt-bn-copyedit}. Post-correction reduced formatting issues and spelling errors; additionally, approximately 10\% of question--option pairs were manually reviewed for quality assurance.

\begin{figure}[H]
\centering
\small
\begin{tcolorbox}[title=Bengali Copy-Editing Prompt, width=\linewidth]
You are a Bengali copy-editor. Correct spelling and grammar only. Preserve MathML math, numerals and answer keys unchanged. Return just the corrected line.
\end{tcolorbox}
\caption{Prompt used for Bengali copy-editing, formatted consistently with our evaluation prompt boxes.}
\label{fig:prompt-bn-copyedit}
\end{figure}

\begin{figure}[ht]
  \centering
  \setlength{\tabcolsep}{1.5pt}       
  \setlength{\arrayrulewidth}{0.3pt}  
  \begin{tabular}{|m{0.315\columnwidth}|m{0.315\columnwidth}|m{0.315\columnwidth}|}
    \hline
    \includegraphics[width=\linewidth]{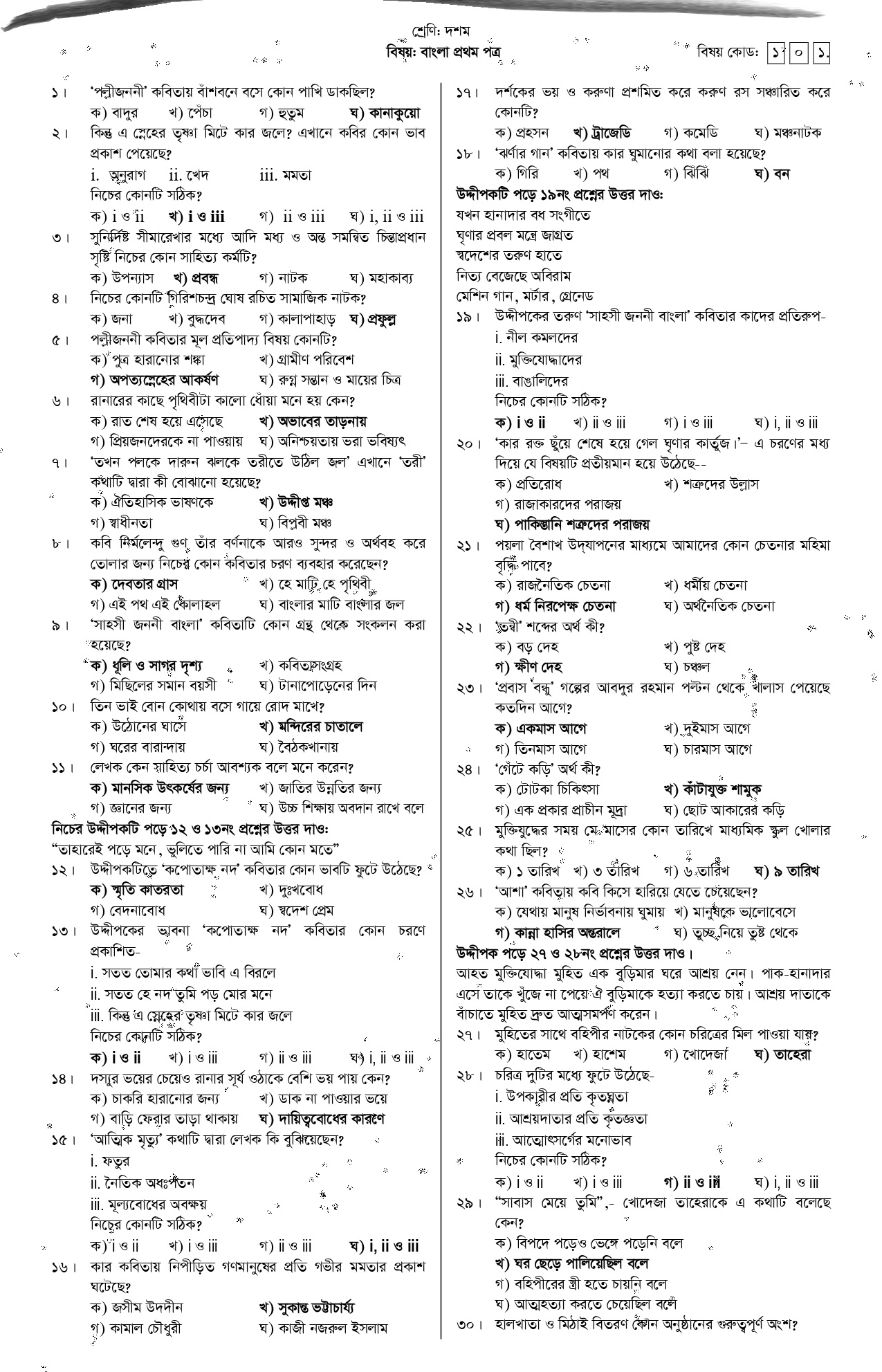} &
    \includegraphics[width=\linewidth]{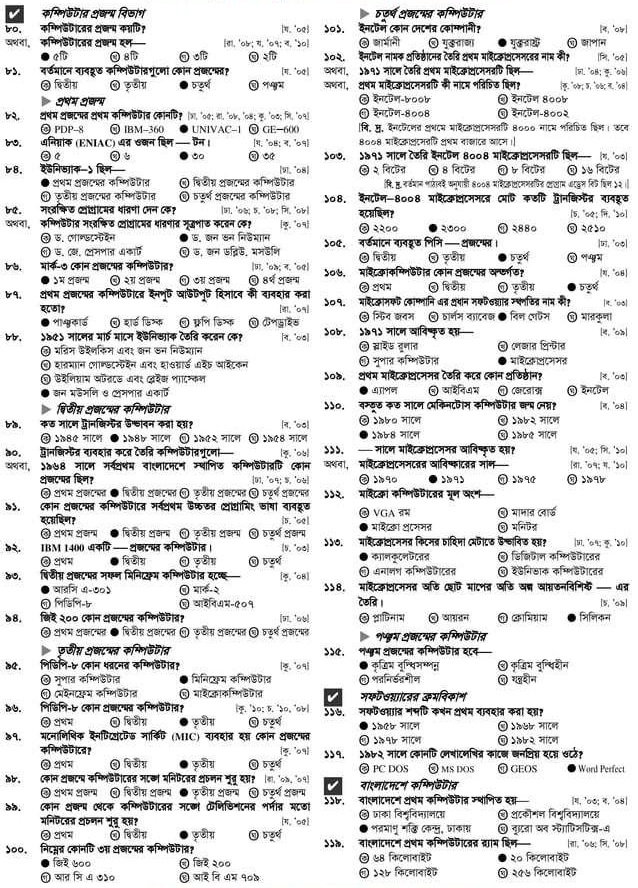} &
    \includegraphics[width=\linewidth]{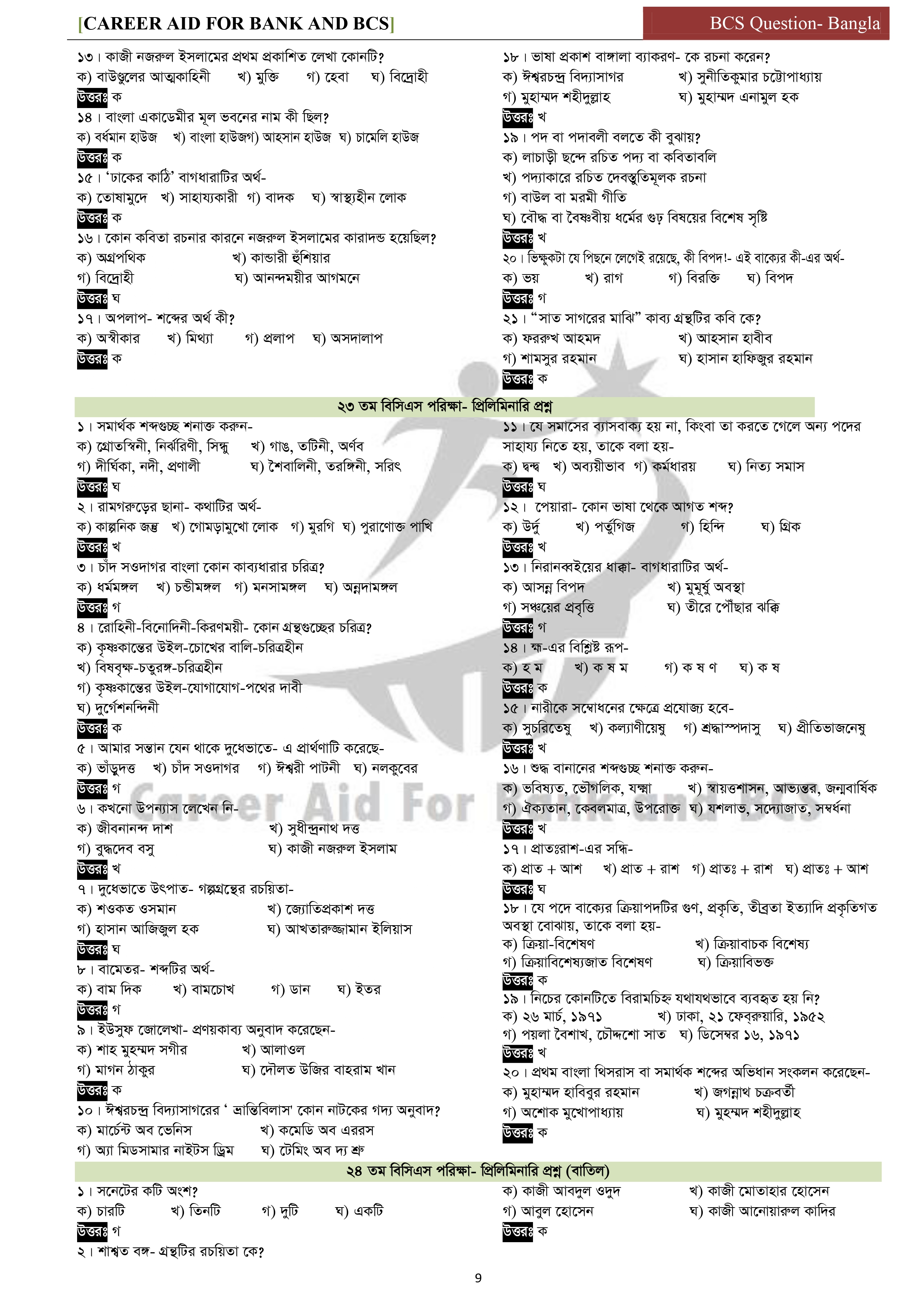} \\
    \hline
  \end{tabular}
  \caption{Sample scanned pages of Bengali multiple-choice questions collected from academic and preparatory guidebooks.} \label{fig:three_with_borders}
\end{figure}

\section{Duplicate-Question Detection and De-duplication}
\label{app:duplicate}
Each question–option pair was embedded into a 1536-dimensional semantic space using the \textit{text-embedding-3-small}\footnote{\href{https://platform.openai.com/docs/models/text-embedding-3-small}{https://platform.openai.com/docs/models/text-embedding-3-small}} model and approximate nearest-neighbor (ANN) search with the angular metric was used to identify semantically similar items. For each question $q_i$, the top-$k$ neighbors $\{q_j\}$ were retrieved and similarity was computed as \autoref{eq:asas}.
\begin{equation} \label{eq:asas}
s(q_i, q_j) = 1 - \frac{d(q_i, q_j)}{2}
\end{equation}
In the \autoref{eq:asas}, $d(\cdot,\cdot)$ is the ANN angular distance. Pairs with $s(q_i, q_j) \ge 0.90$ were flagged as duplicates. These pairs formed an undirected graph $G=(V,E)$, whose connected components defined duplicate clusters. One canonical item per cluster was retained according to a deterministic rule, yielding a de-duplicated and semantically balanced benchmark. The algorithm is shown in \autoref{alg:dedup}.


\begin{algorithm}[h]
\caption{Duplicate-Question Detection and De-duplication}
\label{alg:dedup}
\small
\begin{algorithmic}[1]
\Require Dataset $\mathcal{Q}=\{q_1,\dots,q_N\}$; neighbors $k$; similarity threshold $\tau{=}0.90$
\Ensure Deduplicated set $\mathcal{Q}'$
\State Initialize graph $G \gets \varnothing$
\For{each $q_i \in \mathcal{Q}$}
  \State $v_i \gets \textsc{Embed}(q_i) \in \mathbb{R}^{1536}$
\EndFor
\State Build ANN index over $\{v_i\}_{i=1}^N$
\For{each $q_i \in \mathcal{Q}$}
  \State $\mathcal{N} \gets \textsc{TopKNeighbors}(v_i, k)$
  \For{each $q_j \in \mathcal{N}$}
    \State $s \gets 1 - d(v_i, v_j)/2$
    \If{$i \neq j$ \textbf{and} $s \ge \tau$}
      \State Add edge $(i, j)$ to $G$
    \EndIf
  \EndFor
\EndFor
\State Find connected components $\{C_1,\dots,C_m\}$ of $G$
\State For each component $C_\ell$, retain one canonical question and discard the rest
\State \Return $\mathcal{Q}'$
\end{algorithmic}
\normalsize
\end{algorithm}


\section{Training-test Decontamination Details}
\label{app:decontamination}

To more precisely quantify possible training contamination on LLMs, we perform an $n$-gram decontamination analysis between our multiple-choice test set (questions including answer options) and a broad collection of Bengali corpora and pre-training datasets that are publicly documented or known to be used in at least some of the evaluated models. Because around 80\% of \textsc{BnMMLU} is sourced from web-based question banks, this analysis is critical for ruling out benchmark inflation due to memorization.

\paragraph{Preprocessing and $n$-gram extraction.}
For each test question, we apply Unicode NFKC normalization and collapse consecutive whitespace. We then concatenate the question stem with all answer options into a single sequence, tokenize via simple whitespace splitting and extract all contiguous 13-grams (sequences of 13 tokens). We adopt 13-grams following the GPT-3 contamination protocol and subsequent studies, which treat a shared 13-token span between training and evaluation text as a conservative indicator of near-verbatim reuse rather than incidental overlap \citep{10.5555/3495724.3495883, ravaut2025comprehensivesurveycontaminationdetection}.

\paragraph{Corpora and contamination criterion.}
For each candidate corpus, we stream through the training split and compute the set of 13-grams for every document. A test question is marked as contaminated if any of its 13-grams appear in any training document from at least one corpus. This yields both per-corpus contamination rates and an overall contamination flag per question.

\paragraph{Results.}
\autoref{tab:decontam-breakdown} reports the per-corpus contamination statistics. For Pralekha, Bangla-Instruct, Bangla-TextBook, IndicCorp, OSCAR, CC100, and TituLM, fewer than $0.1\%$ of test questions contain any overlapping 13-gram.

\section{Prompting Styles}
\paragraph{Direct (No-CoT).}
Models are prompted \emph{without} any instruction to explain or “think step by step.” The prompt states the task and requests only the final answer. No intermediate reasoning cues or scaffolded hints are provided.

\paragraph{Chain-of-Thought (CoT).}
Models are explicitly invited to reason before giving the final answer. Prompts include a short instruction to first provide reasoning and then the answer. For comparability, the answer must be clearly marked at the end.


\section{Context Regimes}
\paragraph{Zero-shot (0-shot).}
No exemplars are given; the model receives only the task instruction and the test item (plus CoT cue when applicable). 0 Shot Direct and CoT examples prompts are given in \autoref{fig:0shot-direct} and \autoref{fig:0shot-cot}.

\paragraph{Five-shot (5-shot).}
We supply five worked exemplars per \emph{subdomain}. Each exemplar contains the question, a correct answer, and (for CoT) a concise reasoning trace. The same five exemplars are reused for all test items within that subdomain to ensure consistency. 5-Shot Direct and CoT example prompts are given in \autoref{fig:5shot-direct} and \autoref{fig:5shot-cot}.
\section{Reasoning Configurations}
\paragraph{Reasoning-On (internal).}
For models that has an internal ``reasoning'' or ``thinking'' mode, we additionally evaluate a \emph{Reasoning-On} configuration in the 0-shot setting. Instead of injecting explicit CoT exemplars into the prompt, we enable the provider’s built-in reasoning controls so that the model generates and uses its internal reasoning traces.

\paragraph{Non-Reasoning-On (internal).}
We run reasoning-capable models with their explicit reasoning or ``thinking'' features disabled, using each provider’s control parameter (e.g., \texttt{reasoning\_effort}, \texttt{thinking\_budget}) to suppress chain-of-thought tokens and approximate a standard non-reasoning chat setting. For \textsc{GPT-5-Mini} specifically, we set \texttt{reasoning\_effort = minimal} and \texttt{verbosity = low}; according to OpenAI’s documentation and third-party guidance, this configuration greatly reduces visible reasoning tokens.

\begin{figure}[h]
\centering
\footnotesize
\captionsetup[subfigure]{labelformat=simple, labelsep=space}
\renewcommand\thesubfigure{(\alph{subfigure})}

\begin{subfigure}[t]{0.48\linewidth}
\begin{tcolorbox}[title=0-shot Direct Eval Prompt, width=\linewidth]
\textbf{Question}: \{Question\}\\[2pt]
A.~\{A\} B.~\{B\}\\
C.~\{C\} D.~\{D\}\\
Answer. Output in JSON: \texttt{\{"answer":"A/B/C/D"\}}
\end{tcolorbox}
\caption{} \label{fig:0shot-direct}
\end{subfigure}
\hfill
\begin{subfigure}[t]{0.48\linewidth}
\begin{tcolorbox}[title=0-shot CoT Eval Prompt, width=\linewidth]
\textbf{Question}: \{Question\}\\[2pt]
A.~\{A\} B.~\{B\}\\
C.~\{C\} D.~\{D\}\\
Think step by step. After thinking, provide the final answer in JSON:
\texttt{\{"answer":"A/B/C/D"\}}
\end{tcolorbox}
\caption{}\label{fig:0shot-cot}
\end{subfigure}

\vspace{0.8em}

\begin{subfigure}[t]{0.48\linewidth}
\begin{tcolorbox}[title=5-shot Direct Eval Prompt, width=\linewidth]
\textbf{Example 1}\\
Question: \{Q1\}\\
A.~\{A1\} B.~\{B1\}\\
C.~\{C1\} D.~\{D1\}\\
Answer: \{A/B/C/D\}\\[1pt]
\textbf{Example 2}\\
\ldots\\[6pt]
\textbf{Now solve:}\\
Question: \{Question\}\\
A.~\{A\} B.~\{B\}\\
C.~\{C\} D.~\{D\}\\
Answer: \texttt{\{"answer":"A/B/C/D"\}}
\end{tcolorbox}
\caption{}\label{fig:5shot-direct}
\end{subfigure}
\hfill
\begin{subfigure}[t]{0.48\linewidth}
\begin{tcolorbox}[title=5-shot CoT Eval Prompt, width=\linewidth]
\textbf{Example 1}\\
Question: \{Q1\}\\
A.~\{A1\} B.~\{B1\}\\
C.~\{C1\} D.~\{D1\}\\
Reasoning: \{Reasoning\}\\
Answer: \{A/B/C/D\}\\[1pt]
\textbf{Example 2}\\
\ldots\\[6pt]
\textbf{Now solve:}\\
Question: \{Question\}\\
A.~\{A\} B.~\{B\}\\
C.~\{C\} D.~\{D\}\\
Think step by step. Answer: \texttt{\{"answer":"A/B/C/D"\}}
\end{tcolorbox}
\caption{}\label{fig:5shot-cot}
\end{subfigure}

\caption{Prompts used in our evaluation: 0-shot (Direct, CoT) and 5-shot (Direct, CoT).}
\label{fig:prompts-grid}
\end{figure}

\begin{figure}[H]
\centering
\footnotesize
\begin{tcolorbox}[title=CoT Elicitation Prompt, width=0.95\linewidth]
The following is a MCQ question \{subject\}. Produce a detailed step-by-step explanation in English. Answer and reasoning: \vspace{-2pt}\begin{verbatim}{"answer":"A/B/C/D", "reasoning":"..."}\end{verbatim}
\vspace{-2pt}
Question: \{Question\}\\
A. \{A\} B. \{B\} C. \{C\} D. \{D\}
\end{tcolorbox}
\caption{Prompt used to create the CoT selected questions' reasonings for CoT evaluation.}
\label{fig:cot-example}
\end{figure}

\section{Scaling \& Family Effects}
\begin{figure}[H]
    \centering
    \includegraphics[width=\linewidth]{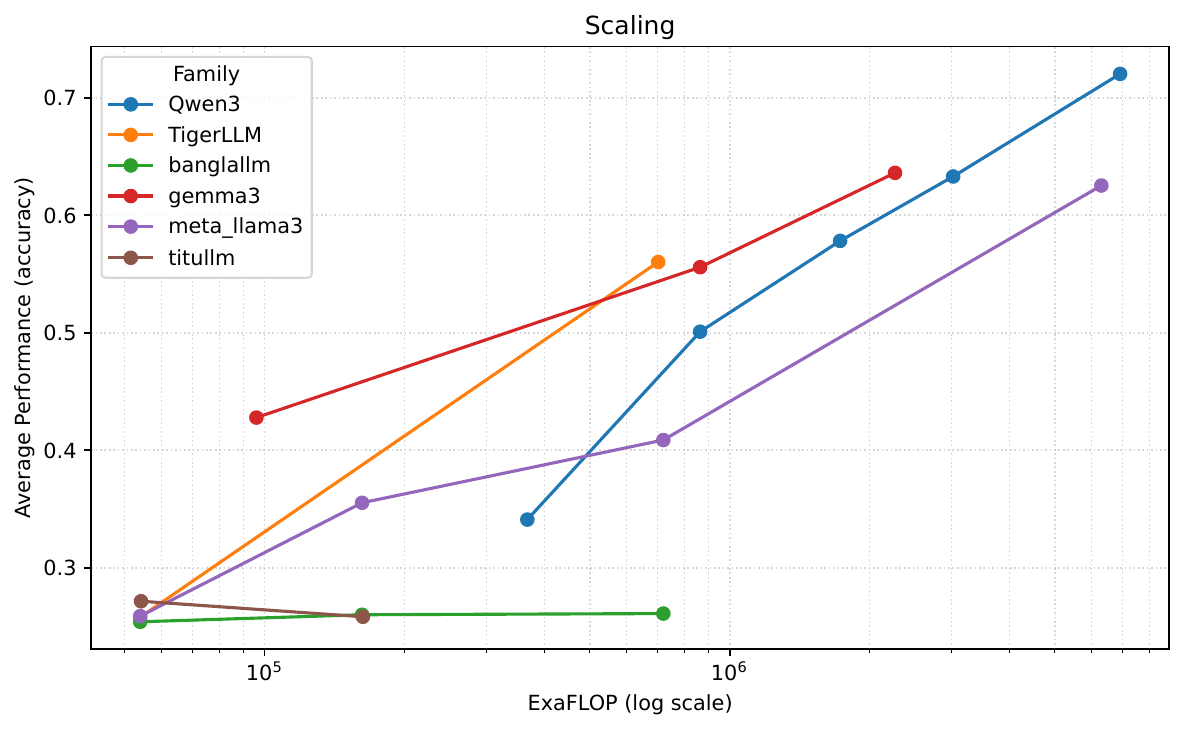}
    \caption{Average accuracy versus estimated training compute (ExaFLOP; log scale). ExaFLOP is estimated as $6 \times \mathrm{params}_{\mathrm{B}} \times \mathrm{train\_tokens}_{\mathrm{B}}$ (both in billions), following Scaling Laws \cite{DBLP:journals/corr/abs-2001-08361}. Accuracy is the per-model mean from 0-shot Direct (Non-Reasoning).}
    \label{fig:scaling}
\end{figure}

Across families, the scaling plot in \autoref{fig:scaling} (ExaFLOP vs.\ average accuracy) shows mostly monotonic “family ladders”: larger, higher-compute checkpoints outperform smaller ones, but gains taper as compute rises. At comparable compute, noticeable cross-family gaps persist-pointing to differences in data curation, pretraining mix, and instruction-tuning rather than scale alone. Bengali-centric families are competitive in the low–mid compute band yet appear to plateau earlier than the largest bilingual/global families.

\section{Sequence-Length Robustness}
\paragraph{Setup.}
To quantify how reliably each model handles longer contexts, we measure error rates as a function of question length.  Let $q$ denote a question, $m$ a model
and $|q|$ the number of characters in $q$.
The procedure is formalised in
Equations \ref{eq:length}–\ref{eq:overall-er}.

\begin{align}
&\text{Length}(q) = |q| \label{eq:length} \\[2pt]
&\text{Bin}(|q|) = \text{bin}_i \quad\text{if}\quad \text{bin}_{i-1} < |q| \le \text{bin}_{i} \label{eq:binning} \\[2pt]
&E(q,m) = 
  \begin{cases}
    0, & \text{if $m$ answers $q$ correctly} \\
    1, & \text{otherwise}
  \end{cases} \label{eq:error} \\[2pt]
&A(m,\text{bin}_i) = 1 - \frac{\sum_{q:\, \text{Bin}(|q|) = \text{bin}_i} E(q,m)}{n_i} \label{eq:accuracy} \\[2pt]
&ER(m,\text{bin}_i) = 1 - A(m,\text{bin}_i) \label{eq:bin-er} \\[2pt]
&ER(m) = \frac{\sum_{q} E(q,m)}{N} \label{eq:overall-er}
\end{align}


The length bins are fixed at $\{0,20,40,60,80,100\}$, $n_i$ is the
number of questions falling in $\text{bin}_i$ and $N$ is the total number of questions. yields the length-specific error rate.

\begin{table}[h]
  \centering
  \small
  \begin{tabular}{cc @{}}
    \toprule
    \textbf{Quadrant}                  & \textbf{Condition}                                      \\
    \midrule
    \midrule
    \textit{Difficult \& Inconsistent} & $\mathrm{Avg}_s < \mu_x \,\land\, \mathrm{SD}_s > \mu_y$ \\
    \textit{Easy \& Inconsistent}      & $\mathrm{Avg}_s > \mu_x \,\land\, \mathrm{SD}_s > \mu_y$ \\
    \textit{Difficult \& Consistent}   & $\mathrm{Avg}_s < \mu_x \,\land\, \mathrm{SD}_s < \mu_y$ \\
    \textit{Easy \& Consistent}        & $\mathrm{Avg}_s > \mu_x \,\land\, \mathrm{SD}_s < \mu_y$ \\
    \bottomrule
  \end{tabular}
  \caption{Quadrant definitions for subject difficulty versus consistency based on average accuracy ($\mathrm{Avg}_s$) and standard deviation ($\mathrm{SD}_s$) thresholds $\mu_x$ and $\mu_y$.}
  \label{tab:quaddddd}
\end{table}

\section{Subject-Specific Failure Modes}

\paragraph{Setup.}
To better understand how language models perform across different subjects, we analyze their subject-wise accuracy and variability. This analysis identifies which subjects are consistently easy or difficult for most models and which ones reveal significant disagreement.

\label{sec:setup}

Here, $\text{Accuracy}_{s,i}$ denote the accuracy of model $i$ on subject
$s$ and $N$ the number of evaluated models.
\begin{align}
&\text{Avg}_s = \frac{1}{N}\sum_{i=1}^{N}\text{Accuracy}_{s,i}
\\
&\text{SD}_s = \sqrt{\frac{1}{N}\sum_{i=1}^{N}\left(
  \text{Accuracy}_{s,i} - \text{Avg}_s
\right)^2}
\end{align}

$\text{Avg}_s$ serves as the X-coordinate and $\text{SD}_s$ as the Y-coordinate in the \emph{Subject Difficulty vs.\ Consistency} plot in \autoref{fig:fig1}.

\section{Compute Resources}
Open-weight models were evaluated on an internal compute node with \textit{1 $\times$ NVIDIA RTX A6000} (48GB) for smaller models and \textit{2 $\times$ NVIDIA RTX PRO 6000} (192GB) for larger models. Open-weight models were executed at their highest native precision; typically \texttt{bfloat16}/\texttt{float16}-with no quantization. Proprietary models were accessed via their official APIs using identical prompts and decoding parameters to ensure comparability across systems.

\begin{figure}[H]
\centering
\footnotesize
\begin{tcolorbox}[title=Bengali (Literature), width=0.98\linewidth]

\textbf{Question (Bengali).} '{\bng iTikya} {\bng thaka{I}} {\bng crm} {\bng sar/thkta} {\bng ny}.'- {\bng mn/tbYiT} {\bng ekan:/} {\bng rcna} {\bng ethek} {\bng en{O}ya}?\\[2pt]
\textit{\textbf{Translation:} 'Merely surviving is not the ultimate success.' - From which composition is this comment taken?}\\[2pt]
\textbf{Options.}
A.~{\bng {oi}Hmn/tii} \textit{(Haimanti)} \\
\textbf{B.~{\bng \sh{iblasii}} \textit{(Bilasi)}} \\
C.~{\bng Ar/dhaNG/gii} \textit{(Ardhangi)} \\
D.~{\bng eJoubenr} {\bng gan}. \textit{(The Song of Youth)}

\medskip
\textbf{\textit{RIGHT - 0-shot-cot-nonreasoning $\to$ B}}\\
\textit{Model: gemini-2.5-flash}

\medskip
\textbf{\textit{WRONG - 5-shot-cot-nonreasoning $\to$ D}}\\
\textit{Model: gemini-2.5-flash}
\end{tcolorbox}
\caption{Exemplar-induced overthinking: 5-shot CoT gravitates to a salient title (D), while 0-shot selects the correct source (B).}

\label{fig:bilashi}
\end{figure}

\begin{figure}[H]
\centering
\footnotesize
\begin{tcolorbox}[title=Mechanics, width=0.98\linewidth]

\textbf{Question (Bengali).} {\bng prRithbiir} {\bng ekn/dR} {\bng Het} {\bng bhuuprRSh/Th} {\bng pr/Jn/t} {\bng duurt/b} {\bng brRid/dhr} {\bng saeth} {\bng saeth} 'g'-{\bng Er} {\bng man}-\\[2pt]
\textit{\textbf{Translation:} From the center of the Earth to the surface, with increasing distance, the value of 'g' -}\\[2pt]
\textbf{Options.}
A.~{\bng smanupaet} {\bng HRas} {\bng pay} \textit{(decreases proportionally)} \\
\textbf{B.~{\bng \sh{smanupaet}} {\bng \sh{brRid/dh}} {\bng \sh{pay}} \textit{(increases proportionally)}} \\
C.~{\bng bYs/tanupaet} {\bng HRas} {\bng pay} \textit{(decreases inversely)} \\
D.~{\bng bYs/tanupaet} {\bng brRid/dh} {\bng pay} \textit{(increases inversely)}

\medskip
\textbf{\textit{WRONG - 5-shot CoT (non-reasoning) $\to$ C}}\\
\textbf{\textit{MODEL REASONING:}} \dots $g \propto {1}/{r^{2}}$ \dots as $r$ increases, $g$ decreases \dots \textbf{“{\bng bYs/tanupaet} {\bng HRas} {\bng pay}”} matches inverse square \dots \verb|{"answer":"C"}| \dots\\
\textit{Context:} CoT assumes \textbf{outside-Earth regime} and never switches to the inside-Earth model.

\medskip
\textbf{\textit{RIGHT - 0-shot Reasoning-On $\to$ B}}\\
\textbf{\textit{MODEL REASONING:}} \dots \textbf{inside Earth} ($r<R$), $g \propto r$ \dots from center $\to$ surface, \textbf{$g$ increases proportionally} \dots \textbf{Therefore B} \dots\\
\textit{Context:} Reasoning-on selects the \textbf{correct regime} (interior of a uniform sphere) and \textbf{maps wording $\to$ option}.
\end{tcolorbox}
\caption{Physics: Reasoning-On toggles to the inside-sphere linear model ($g\!\propto\!r$), correcting CoT’s \textbf{inverse-square overgeneralization}.}
\label{fig:physics}
\end{figure}

\begin{figure}[H]
\centering
\footnotesize
\begin{tcolorbox}[title=Business Strategy \& Management, width=0.98\linewidth]

\textbf{Question (Bengali).} {\bng EkiT} {\bng baiNijYk} {\bng bYaNNGk} {\bng sNJ/cJii} {\bng iHsab} {\bng khuel} {\bng tuim} \textbf{{\bng sp/taeHr}} {\bng sr/baidhk} {\bng ktbar} {\bng Takar} {\bng tulet} {\bng paera}?\\[2pt]
\textit{\textbf{Translation:} By opening a savings account in a commercial bank, how many times at most can you withdraw money \textbf{per week}?}\\[2pt]
\textbf{Options.}
\textbf{A.~{\bng \sh{2}} {\bng \sh{bar}} \textit{(2 times)}}\\
B.~{\bng 3} {\bng bar} \textit{(3 times)}\\
C.~{\bng 4} {\bng bar} \textit{(4 times)}\\
D.~{\bng Jtbar} {\bng {I}c/cha} \textit{(As many times as desired)}

\medskip
\textbf{\textit{WRONG - 5-shot CoT (non-reasoning) $\to$ C}}\\
\textbf{\textit{MODEL REASONING:}} \dots savings accounts have a limit \dots up to \textbf{4 withdrawals} \dots Therefore \textbf{4} \dots \verb|{"answer":"C"}| \dots\\
\textit{Context:} CoT defaults to \textbf{monthly limits} learned from exemplars; misreads the \textbf{“per week”} cue.

\medskip
\textbf{\textit{RIGHT - 0-shot Reasoning-On $\to$ A}}\\
\textbf{\textit{MODEL REASONING:}} \dots interpret as \textbf{“per week”} \dots savings accounts limited withdrawals \dots \textbf{commonly 2 per week} \dots \textbf{Therefore A} \dots\\
\textit{Context:} Reasoning-on \textbf{re-parses the Bengali phrase}, aligns to the \textbf{weekly rule}, then \textbf{option-checks}.
\end{tcolorbox}
\caption{Strategy: Reasoning-On corrects \textbf{temporal granularity} (weekly vs.\ monthly), avoiding CoT’s \textbf{exemplar-driven heuristic}.}
\label{fig:gran}
\end{figure}



\begin{figure*}[h]
    \centering
    \includegraphics[width=\linewidth]{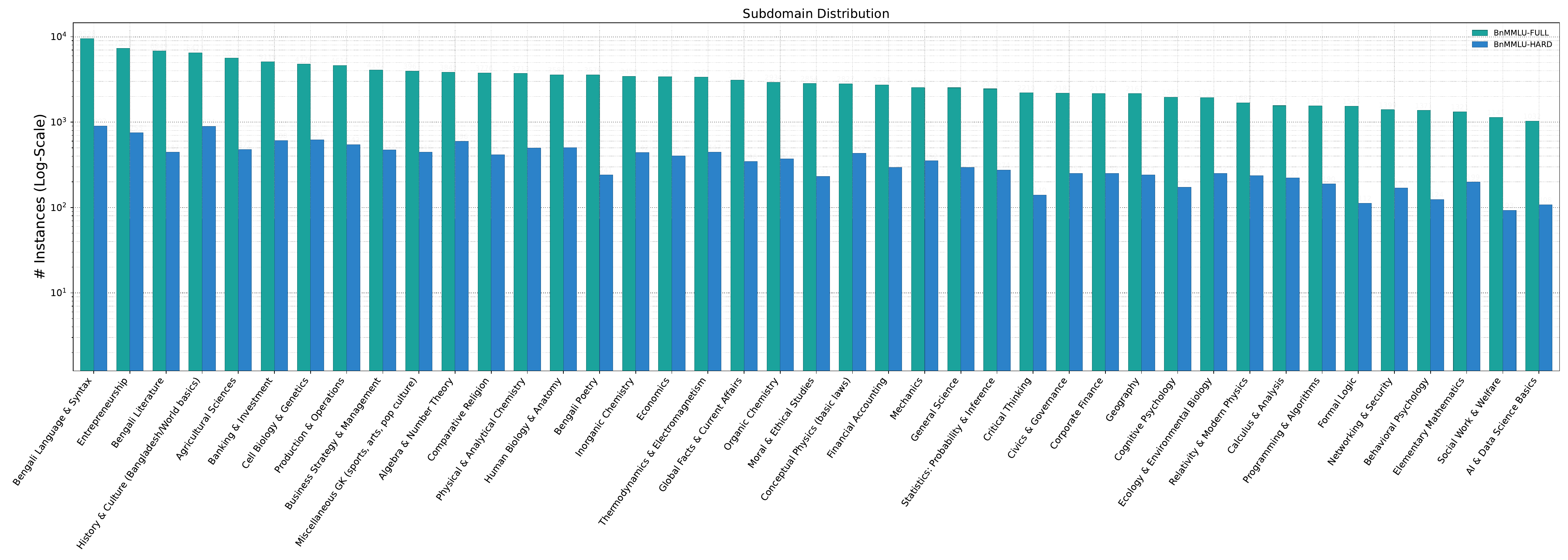}
    \caption{Subject-wise counts for \textbf{BnMMLU-FULL} and \textbf{BnMMLU-HARD}.}
    \label{fig:placeholder}
\end{figure*}

\begin{table*}[h]
\centering
\footnotesize
\begin{threeparttable}
\setlength{\tabcolsep}{6pt}\renewcommand{\arraystretch}{1.15}
\begin{tabular}{lrrrr}
\toprule
\textbf{Corpus (dataset / split)} & \textbf{\# Ref. Exp.} & \textbf{\#Cont. Qs.} & \textbf{\# Cont. Qs. (\%)} & \textbf{\# Unq. 13-g} \\
\midrule
\midrule
Pralekha (ben) \citep{suryanarayanan2025pralekhacrosslingualdocumentalignment}          & 95{,}813      & 0   & 0.00 & 0   \\
Pralekha (eng-ben) \citep{suryanarayanan2025pralekhacrosslingualdocumentalignment}         & 86{,}815      & 0   & 0.00 & 0   \\
Pralekha (unal / ben) \citep{suryanarayanan2025pralekhacrosslingualdocumentalignment}        & 47{,}906      & 1   & 0.00 & 2   \\
TituLM Corpus \citep{nahin2025titullmsfamilybanglallms}               & 31{,}225{,}356 & 122 & 0.09 & 239 \\
IndicCorpV2 (asm--Beng) \citep{doddapaneni-etal-2023-towards}  & 1{,}256{,}513 & 0   & 0.00 & 0   \\
IndicCorpV2 (Beng) \citep{doddapaneni-etal-2023-towards}      & 13{,}553{,}516 & 30  & 0.02 & 80  \\
OSCAR (bn) \citep{ortiz-suarez-etal-2020-monolingual}             & 14{,}346{,}126 & 34  & 0.02 & 79  \\
Bangla-Instruct (Instruction) \citep{raihan-zampieri-2025-tigerllm}      & 268{,}145     & 4   & 0.00 & 3   \\
Bangla-Instruct (Response) \citep{raihan-zampieri-2025-tigerllm}     & 329{,}872     & 49  & 0.04 & 219 \\
Bangla-TextBook  \citep{raihan-zampieri-2025-tigerllm}           & 87{,}105      & 48  & 0.03 & 209 \\
CC100 (bn) \citep{wenzek-etal-2020-ccnet}                  & 12{,}427{,}522 & 72 & 0.05 & 141 \\
\bottomrule
\end{tabular}
\caption{13-gram decontamination statistics. A question is marked as contaminated if its normalized text (question plus answer options) shares at least one contiguous 13-gram with any example in the corresponding training corpus.}
\label{tab:decontam-breakdown}
\end{threeparttable}
\end{table*}

\begin{table*}[h]
\centering
\footnotesize
\begin{threeparttable}
\setlength{\tabcolsep}{6pt}\renewcommand{\arraystretch}{1.15}
\begin{tabular}{lccc}
\toprule
\textbf{Model} & \textbf{\# Params} & \textbf{Access} & \textbf{Language} \\
\midrule
\midrule
\multicolumn{4}{c}{\textit{\textbf{English-Centric / Bilingual Instruction-Tuned Models}}} \\
\midrule
\textsc{Llama-3.x-Instruct} \cite{grattafiori2024llama3herdmodels} & 1B, 3B, 8B, 70B & Weights Available & En / Multilingual \\
\textsc{Qwen3} \cite{yang2025qwen3technicalreport} & 1.7B, 4B, 8B, 14B, 32B & Weights Available & En / Zh \\
\textsc{Gemma-3-IT} \cite{gemmateam2025gemma3technicalreport} & 4B, 12B, 27B & Weights Available & En / Multilingual \\
\midrule
\multicolumn{4}{c}{\textit{\textbf{Bengali Pretrained / Instruction-Tuned Models}}} \\
\midrule
\textsc{TituLLM} \cite{nahin-etal-2025-titullms} & 1B, 3B & Weights Available & Bn / En \\
\textsc{TigerLLM-IT} \cite{raihan-zampieri-2025-tigerllm} & 1B, 9B & Weights Available & Bn \\
\textsc{BanglaLlama Instruct} \cite{zehady2025banglallamallamabanglalanguage} & 1B, 3B, 8B & Weights Available & Bn \\
\midrule
\multicolumn{4}{c}{\textit{\textbf{Proprietary Models}}} \\
\midrule
\textsc{GPT-5-Mini}\tnote{*} \cite{openai2025gpt5systemcard} & undisclosed & API & – \\
\textsc{Grok 4 Fast}\tnote{a} & undisclosed & API & – \\
\textsc{Gemini 2.5 Flash \cite{comanici2025gemini25pushingfrontier}} & undisclosed & API & – \\
\textsc{DeepSeek-V3.2-Exp} \cite{deepseekai2024deepseekv32} & 685B & API / Weights Available & – \\
\textsc{Qwen-Plus}\tnote{b} & undisclosed & API & – \\
\bottomrule
\end{tabular}
\begin{tablenotes}
\footnotesize
\item[a]\url{https://x.ai/news/grok-4-fast}
\item[b]\url{https://www.alibabacloud.com/help/en/model-studio/models}
\item[*] It has reasoning capability but cannot be fully disabled and thus, we use minimal reasoning when mentioning no-reasoning. 
\end{tablenotes}
\end{threeparttable}
\caption{Overview of evaluated models, grouped by family.}
\label{tab:model_overview_grouped}
\end{table*}

\begin{table*}[h]
\centering
\footnotesize
\setlength{\tabcolsep}{4pt}
\renewcommand{\arraystretch}{1.2}
\begin{tabular}{lccccc}
\toprule
\textbf{Model} &
\textbf{STEM} &
\textbf{Humanities} &
\textbf{Social Sciences} &
\textbf{Others} &
\textbf{Overall ($\Delta$)} \\
\midrule
\midrule
\multicolumn{6}{c}{\textit{\textbf{English-Centric / Bilingual Instruction-Tuned Models}}} \\
\midrule
\textsc{Llama-3.2-1B-Instruct}      & 25.88 & 27.22 & 27.26 & 26.71 & 26.69 \textcolor{darkgreen}{(+1.69)} \\
\textsc{Llama-3.2-3B-Instruct}      & 35.53 & 34.91 & 41.34 & 40.30 & 37.68 \textcolor{darkgreen}{(+12.68)} \\
\textsc{Llama-3.1-8B-Instruct}      & 40.87 & 39.49 & 47.14 & 46.89 & 43.07 \textcolor{darkgreen}{(+18.07)} \\
\textsc{Llama-3.3-70B-Instruct}     & 62.53 & 52.47 & \underline{65.81} & \underline{68.99} & 61.87 \textcolor{darkgreen}{(+36.87)} \\
\textsc{Qwen3-1.7B}     & 34.10 & 34.58 & 39.70 & 36.00 & 36.25 \textcolor{darkgreen}{(+11.25)} \\
\textsc{Qwen3-4B}       & 50.09 & 40.60 & 48.46 & 47.21 & 47.27 \textcolor{darkgreen}{(+22.27)} \\
\textsc{Qwen3-8B}       & 57.82 & 43.96 & 51.87 & 53.83 & 52.51 \textcolor{darkgreen}{(+27.51)} \\
\textsc{Qwen3-14B}      & 63.30 & 49.37 & 59.23 & 61.13 & 58.76 \textcolor{darkgreen}{(+33.76)} \\
\textsc{Qwen3-32B}      & \underline{72.03} & \underline{53.01} & 65.57 & 66.44 & \underline{65.34 \textcolor{darkgreen}{(+40.34)}} \\
\textsc{Gemma-3-4B-IT}              & 42.78 & 39.46 & 47.82 & 47.95 & 44.07 \textcolor{darkgreen}{(+19.07)} \\
\textsc{Gemma-3-12B-IT}             & 55.58 & 47.50 & 59.13 & 59.82 & 55.26 \textcolor{darkgreen}{(+30.26)} \\
\textsc{Gemma-3-27B-IT}             & 63.61 & 51.43 & 63.90 & 67.44 & 61.27 \textcolor{darkgreen}{(+36.27)} \\
\midrule
\multicolumn{6}{c}{\textit{\textbf{Bengali Pretrained / Instruction-Tuned Models}}} \\
\midrule
\textsc{TituLLM-1B}                 & 27.15 & 27.53 & 28.42 & 28.19 & 27.72 \textcolor{darkgreen}{(+2.72)} \\
\textsc{TituLLM-3B}                 & 25.83 & 27.59 & 27.70 & 26.56 & 26.87 \textcolor{darkgreen}{(+1.87)} \\
\textsc{TigerLLM-1B-IT}             & 25.80 & 27.47 & 27.42 & 26.11 & 26.73 \textcolor{darkgreen}{(+1.73)} \\
\textsc{TigerLLM-9B-IT}             & \underline{56.02} & \underline{47.85} & \underline{59.48} & \underline{61.29} & \underline{55.70 \textcolor{darkgreen}{(+30.70)}} \\
\textsc{BanglaLLaMA-3.2-1B-Instruct} & 25.40 & 27.40 & 27.07 & 25.93 & 26.40 \textcolor{darkgreen}{(+1.40)} \\
\textsc{BanglaLLaMA-3.2-3B-Instruct} & 26.00 & 27.56 & 27.44 & 26.26 & 26.82 \textcolor{darkgreen}{(+1.82)} \\
\textsc{BanglaLLaMA-3.1-8B-Instruct} & 26.10 & 27.56 & 27.58 & 26.52 & 26.95 \textcolor{darkgreen}{(+1.95)} \\
\midrule
\multicolumn{6}{c}{\textit{\textbf{Proprietary Models}}} \\
\midrule
\textsc{GPT-5-Mini}         & 48.25 & 43.96 & 55.00 & 55.78 & 50.09 \textcolor{darkgreen}{(+25.09)} \\
\textsc{Grok 4 Fast}    & 61.98 & 51.63 & 64.02 & 67.60 & 60.82 \textcolor{darkgreen}{(+35.82)} \\
\textsc{Gemini 2.5 Flash} & 72.38 & \textbf{62.32} & \textbf{71.08} & \textbf{73.85} & \textbf{69.85 \textcolor{darkgreen}{(+44.85)}} \\
\textsc{DeepSeek-V3.2-Exp} & 72.72 & 58.62 & 70.06 & 73.84 & 68.82 \textcolor{darkgreen}{(+43.82)} \\
\textsc{Qwen-Plus}      & \textbf{73.49} & 56.15 & 66.89 & 70.17 & 67.29 \textcolor{darkgreen}{(+42.29)} \\
\bottomrule
\end{tabular}
\caption{Average accuracy (\%) of models on the \textbf{BnMMLU-FULL} benchmark under 0-shot Direct (Non-Reasoning) evaluation. Bold marks the highest overall score; underlines denote the best model within each category. ($\Delta$) in overall is compared with random baseline (25\%).}
\label{tab:bnmmlu_full}
\end{table*}

\begin{table*}[h]
\centering
\footnotesize
\setlength{\tabcolsep}{4pt}
\renewcommand{\arraystretch}{1.2}
\begin{tabular}{lcccc}
\toprule
\textbf{Model} &
\makecell{\textbf{0-shot Direct}\\\textbf{(Non-Reasoning)}} &
\makecell{\textbf{0-shot CoT ($\Delta$)}\\\textbf{(Non-Reasoning)}} &
\makecell{\textbf{5-shot Direct}\\\textbf{(Non-Reasoning)}} &
\makecell{\textbf{5-shot CoT ($\Delta$)}\\\textbf{(Non-Reasoning)}} \\
\midrule
\midrule
\multicolumn{5}{c}{\textit{\textbf{English-Centric / Bilingual Instruction-Tuned Models}}} \\
\midrule
\textsc{Llama-3.2-3B-Instruct}            & 19.95 & 18.33 \textcolor{red}{(-1.62)} & 22.16 & 23.25 \textcolor{darkgreen}{(+1.09)} \\
\textsc{Llama-3.1-8B-Instruct}            & 19.14 & 22.91 \textcolor{darkgreen}{(+3.77)} & 21.99 & 22.62 \textcolor{darkgreen}{(+0.63)} \\
\textsc{Llama-3.3-70B-Instruct}           & 23.78 & 35.17 \textcolor{darkgreen}{(+11.39)} & 31.15 & \underline{37.50 \textcolor{darkgreen}{(+6.35)}} \\
\textsc{Qwen3-1.7B}                       & 14.53 & 21.67 \textcolor{darkgreen}{(+7.14)} & 23.27 & 23.55 \textcolor{darkgreen}{(+0.28)} \\
\textsc{Qwen3-4B}                         & 12.26 & 19.09 \textcolor{darkgreen}{(+6.83)} & 26.46 & 29.74 \textcolor{darkgreen}{(+3.28)} \\
\textsc{Qwen3-8B}                         & 21.59 & 20.91 \textcolor{red}{(-0.68)} & 29.14 & 28.39 \textcolor{red}{(-0.75)} \\
\textsc{Qwen3-14B}                        & 14.67 & 14.32 \textcolor{red}{(-0.35)} & 18.35 & 16.88 \textcolor{red}{(-1.47)} \\
\textsc{Qwen3-32B}                        & \underline{25.52} & 28.63 \textcolor{darkgreen}{(+3.11)} & 34.63 & 31.19 \textcolor{red}{(-3.44)} \\
\textsc{Gemma-3-4B-IT}                    & 14.85 & 15.72 \textcolor{darkgreen}{(+0.87)} & 16.78 & 19.51 \textcolor{darkgreen}{(+2.73)} \\
\textsc{Gemma-3-12B-IT}                   & 10.54 & 14.55 \textcolor{darkgreen}{(+4.01)} & 18.50 & 23.52 \textcolor{darkgreen}{(+5.02)} \\
\textsc{Gemma-3-27B-IT}                   & 14.72 & \underline{37.59 \textcolor{darkgreen}{(+22.87)}} & \underline{35.65} & 34.65 \textcolor{red}{(-1.00)} \\
\midrule
\multicolumn{5}{c}{\textit{\textbf{Bengali Pretrained / Instruction-Tuned Models}}} \\
\midrule
\textsc{TigerLLM-9B-IT}                   & \underline{11.01} & \underline{16.78 \textcolor{darkgreen}{(+5.77)}} & \underline{18.44} & \underline{23.32 \textcolor{darkgreen}{(+4.88)}} \\
\midrule
\multicolumn{5}{c}{\textit{\textbf{Proprietary Models}}} \\
\midrule
\textsc{GPT-5-Mini}                       & 14.13 & 19.12 \textcolor{darkgreen}{(+4.99)} & 19.66 & 18.63 \textcolor{red}{(-1.03)} \\
\textsc{Grok 4 Fast}                      & 20.94 & 20.89 \textcolor{red}{(-0.05)} & 44.06 & 51.12 \textcolor{darkgreen}{(+7.06)} \\
\textsc{Gemini 2.5 Flash}            & \textbf{34.46} & 45.38 \textcolor{darkgreen}{(+10.92)} & 51.62 & 61.48 \textcolor{darkgreen}{(+9.86)} \\
\textsc{DeepSeek-V3.2-Exp}                & 29.89 & \textbf{59.04 \textcolor{darkgreen}{(+29.15)}} & \textbf{58.83} & \textbf{64.53 \textcolor{darkgreen}{(+5.70)}} \\
\textsc{Qwen-Plus}                        & 32.47 & 58.74 \textcolor{darkgreen}{(+26.27)} & 57.40 & 55.09 \textcolor{red}{(-2.31)} \\
\bottomrule
\end{tabular}
\caption{Accuracy (\%) on \textbf{BnMMLU-HARD}. \(\Delta\) is computed as \(\text{CoT} - \text{Direct}\) at the \emph{same shot} (0-shot or 5-shot). \textbf{Bold} marks the global best per column; \underline{underline} marks the best \emph{within each category} per column.}

\label{tab:bnmmlu_3col_subdelta__}
\end{table*}

\begin{table*}[ht]
  \caption{Overview of subject domains and tested concepts in BnMMLU.}
  \label{tab:mmlu_subjects}
  \centering
  \footnotesize
  \setlength{\tabcolsep}{4pt}
  \renewcommand{\arraystretch}{1.3}
  \begin{tabular*}{\textwidth}{@{\extracolsep{\fill}}l l p{0.48\textwidth} l@{}}
    \toprule
    \textbf{SL} & \textbf{Subject Name} & \textbf{Tested Concepts} & \textbf{Supercategory} \\
    \midrule
     1  & Elementary Mathematics                   & Arithmetic, Fractions, Ratios, Basic Problem Solving\dots      & STEM \\
     2  & Algebra \& Number Theory                 & Equations, Functions, Prime Numbers, Theorems\dots             & STEM \\
     3  & Calculus \& Analysis                     & Differentiation, Integration, Sequences, Series\dots           & STEM \\
     4  & Statistics: Probability \& Inference     & Descriptive Statistics, Probability, Hypothesis Testing\dots   & STEM \\
     5  & Mechanics                                & Dynamics, Statics, Kinematics, Laws of Motion\dots             & STEM \\
     6  & Conceptual Physics (basic laws)          & Motion, Forces, Energy, Newtonian Principles\dots              & STEM \\
     7  & Thermodynamics \& Electromagnetism       & Laws of Thermodynamics, Heat Transfer, Electricity\dots        & STEM \\
     8  & Relativity \& Modern Physics             & Einstein’s Theories, Quantum Concepts, Atomic Models\dots      & STEM \\
     9  & Physical \& Analytical Chemistry         & Stoichiometry, Molecular Structure, Spectroscopy\dots          & STEM \\
    10  & Inorganic Chemistry                      & Periodic Table, Coordination Compounds\dots                    & STEM \\
    11  & Organic Chemistry                        & Hydrocarbons, Functional Groups, Reactions\dots    & STEM \\
    12  & Cell Biology \& Genetics                 & Cell Structure, DNA/RNA, Inheritance, Evolution\dots           & STEM \\
    13  & Human Biology \& Anatomy                 & Organ Systems, Physiology, Human Genetics\dots                 & STEM \\
    14  & Ecology \& Environmental Biology         & Ecosystems, Biodiversity, Conservation, Sustainability\dots    & STEM \\
    15  & Agri Sciences                    & Agronomy, Crop, Soil Management, Agribusiness\dots     & STEM \\
    16  & Networking \& Security                   & Internet Protocols, Cybersecurity, Encryption, Firewalls\dots  & STEM \\
    17  & Programming \& Algorithms                & Python, Logic, Data Structures, Computational Thinking\dots    & STEM \\
    18  & AI \& Data Science Basics                & Machine Learning, Neural Networks, Data Processing\dots        & STEM \\
    19  & General Science                          & Scientific Method, Basic Physics, Chemistry, Biology\dots      & STEM \\
    20  & Bengali Language \& Syntax               & Morphology, Grammar, Sentence Structure, Semantics\dots        & Humanities \\
    21  & Bengali Literature                       & Prose, Poetry, Authors, Literary Devices\dots                  & Humanities \\
    22  & Bengali Poetry                           & Poetic Forms, Symbolism, Meter, Notable Poets\dots             & Humanities \\
    23  & Comparative Religion                     & Theology, World Religions, Ethical Teachings\dots              & Humanities \\
    24  & Moral \& Ethical Studies                 & Ethics, Values, Philosophy, Social Responsibility\dots          & Humanities \\
    25  & Formal Logic                             & Propositional Logic, Proofs, Logical Systems, Paradoxes\dots    & Humanities \\
    26  & Critical Thinking                        & Logic, Reasoning, Argumentation, Analytical Skills\dots         & Humanities \\
    27  & Economics                                & Microeconomics, Macroeconomics, Fiscal Policy, Trade\dots       & Social Sciences \\
    28  & Banking \& Investment                    & Financial Systems, Banking Principles, Securities\dots          & Social Sciences \\
    29  & Financial Accounting                     & Balance Sheets, Cash Flow, Auditing, Cost Analysis\dots         & Social Sciences \\
    30  & Corporate Finance                        & Capital Budgeting, Valuation, Risk Management\dots              & Social Sciences \\
    31  & Business Strategy \& Management          & Strategic Planning, Leadership, Organizational Theory\dots      & Social Sciences \\
    32  & Production \& Operations                 & Process Design, Quality Control, Supply Chain\dots              & Social Sciences \\
    33  & Entrepreneurship                         & Startup Models, Business Planning, Innovation\dots             & Social Sciences \\
    34  & Cognitive Psychology                     & Memory, Perception, Decision-Making, Theories\dots     & Social Sciences \\
    35  & Behavioral Psychology                    & Emotions, Behaviorism, Conditioning, Human Interaction\dots     & Social Sciences \\
    36  & Civics \& Governance                     & Constitution, Rights, Political Systems, Citizenship \dots & Social Sciences \\
    37  & Geography                                & Physical Geography, Climate, Maps, Human Geography\dots         & Social Sciences \\
    38  & History \& Culture                       & Historical Events, Heritage, Civilization, Global Affairs\dots  & Social Sciences \\
    39  & Social Work \& Welfare                   & Social Policy, Community Engagement, Case Studies\dots          & Social Sciences \\
    40  & Miscellaneous GK                         & Global Trivia, Sports Facts, Entertainment, Arts, Media\dots    & Others \\
    41  & Global Facts \& Current Affairs          & International Organizations, Events, World Politics\dots        & Others \\
    \bottomrule
  \end{tabular*}
\end{table*}

\end{document}